\documentclass[10pt,twocolumn,letterpaper]{article}

\usepackage[pagenumbers]{cvpr} 

\definecolor{cvprblue}{rgb}{0.21,0.49,0.74}
\usepackage[pagebackref,breaklinks,colorlinks]{hyperref}

\usepackage{algorithmic}
\usepackage{algorithm}
\usepackage{array}
\usepackage{textcomp}
\usepackage{stfloats}
\usepackage{url}
\usepackage{verbatim}
\usepackage{graphicx}
 \usepackage{wrapfig}

\usepackage{color}
\usepackage{xcolor}
\usepackage{makecell}
\usepackage{colortbl}  
\usepackage{rotating}
\usepackage{footnote}
 \usepackage[accsupp]{axessibility}
\usepackage{multirow} 
\usepackage{booktabs} 
\usepackage{overpic} 
\definecolor{pink}{HTML}{FB3199}

\hypersetup{
    urlcolor=pink,        
    colorlinks=red,
}

\definecolor{lightgrey}{RGB}{211,211,211}

\def\confName{CVPR}
\def\confYear{2025}

\title{Plug-and-Play Versatile Compressed Video Enhancement}
\author{
Huimin Zeng \qquad  Jiacheng Li \qquad  Zhiwei Xiong\thanks{Corresponding author.} \\
University of Science and Technology of China\\
{\tt\small \{zenghuimin, jclee\}@mail.ustc.edu.cn \qquad zwxiong@ustc.edu.cn}
}

\begin{document}
\maketitle
\begin{abstract}
As a widely adopted technique in data transmission, video compression effectively reduces the size of files, making it possible for real-time cloud computing.  However, it comes at the cost of visual quality, posing challenges to the robustness of downstream vision models.  In this work, we present a versatile codec-aware enhancement framework that reuses codec information to adaptively enhance videos under different compression settings, assisting various downstream vision tasks without introducing computation bottleneck.  Specifically, the proposed codec-aware framework consists of a compression-aware adaptation (CAA) network that employs a hierarchical adaptation mechanism to estimate parameters of the frame-wise enhancement network, namely the bitstream-aware enhancement  (BAE) network. The BAE network further leverages temporal and spatial priors embedded in the bitstream to effectively improve the quality of compressed input frames.  Extensive experimental results demonstrate the superior quality enhancement performance of our framework over existing enhancement methods, as well as its versatility in assisting multiple downstream tasks on compressed videos as a plug-and-play module.  Code and models are available at \url{https://huimin-zeng.github.io/PnP-VCVE/}. 
\end{abstract}

\vspace{-8pt}\section{Introduction}
\begin{figure}[t]
    \centering
    \includegraphics[width=1\linewidth]{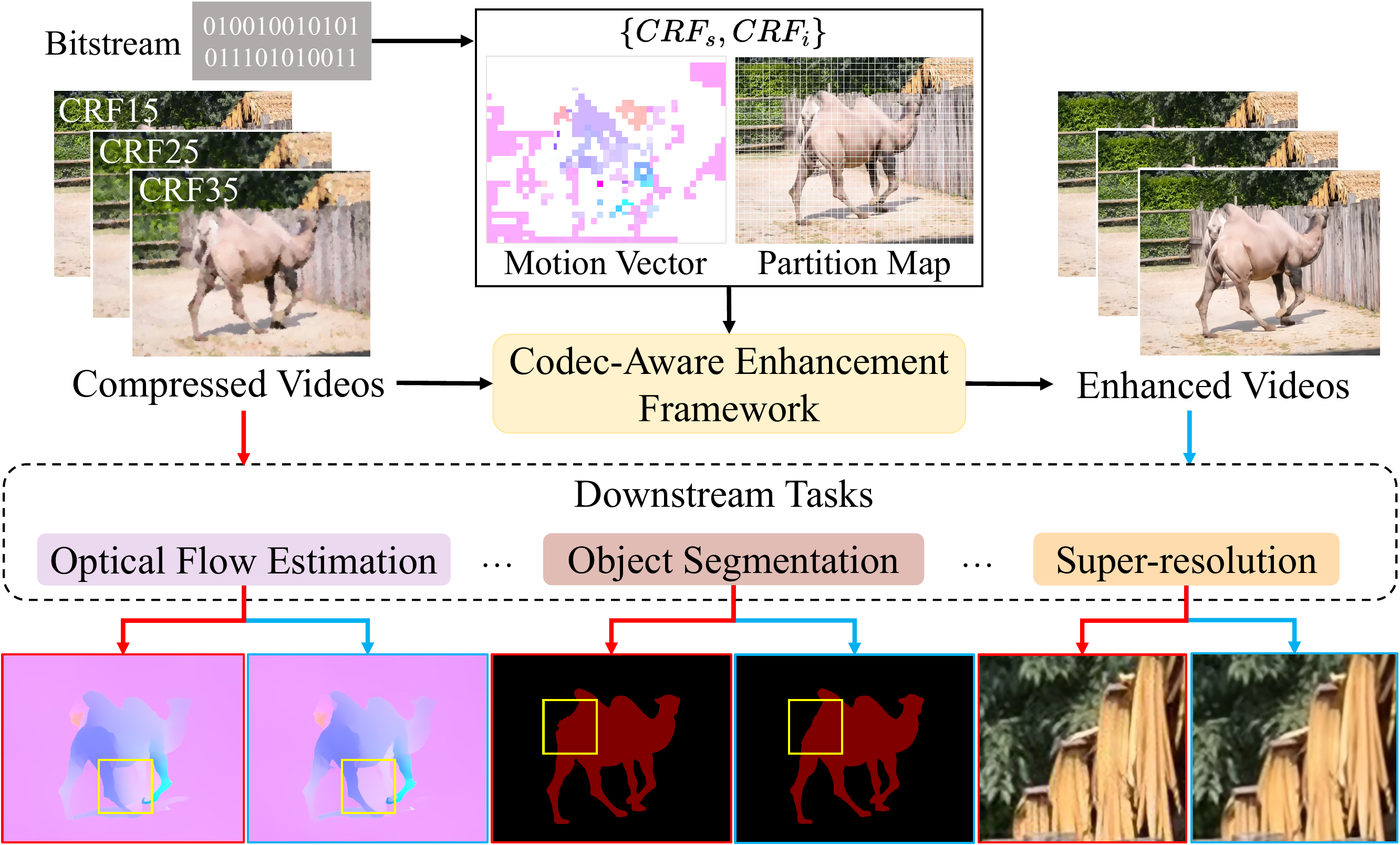} \vspace{-0.2in}
    \caption{The proposed codec-aware enhancement framework reuses codec information to adaptively enhance videos across different compression settings, while assisting in various downstream tasks in a plug-and-play manner.}  \vspace{-0.15in}
    \label{fig:teaser}
\end{figure}

\vspace{-4pt}
With the flower booming of short video platforms,  video has become one of the most popular multimedia formats. In addition to distributing visual content, in practical scenarios (\eg, autonomous driving~\cite{sun2020advanced,siam2018comparative}), it is common to upload the captured videos to the cloud end for further visual analysis and downstream applications (\eg, object detection~\cite{cai2021yolobile,wang2019fast} and segmentation~\cite{siam2021video,feng2020deep}).   However, due to the bandwidth constraint during transmission, these videos are typically compressed with varying levels, resulting in poor visual quality and suboptimal performance in downstream tasks~\cite{zhang2011close,mfqe2} (\eg, inaccurate segmentation boundaries in Fig.~\ref{fig:teaser}). Given the crucial role of videos in data transmission, there is a critical need for a versatile solution to enhance videos of diverse compression levels and effectively support various downstream tasks.

Existing video enhancement methods are hard to respond to these demands. Specifically, to effectively enhance videos of different compression levels, previous methods~\cite{mfqe,mfqe2,stdf,cvcp,chen2020bitstream,lin2022unsupervised} employ separate enhancement models for each compression level, which is inflexible occurring unseen compression levels. Recent approaches~\cite{Ehrlich_2024_WACV,sun2022quality,realbasicvsr} consider this issue as the generalization ability across diverse compression levels, therefore randomly selecting inputs of different compression levels during training.   However, such a training strategy is compression-agnostic and offers limited improvement.   Most importantly, the aforementioned methods focus primarily on improving perceptual quality, neglecting the need to assist in downstream tasks in real-world scenarios.

Based on the mismatch between the versatility demand and existing solutions, here we summarize the following criteria of a favorable solution: 1) adaptively enhance videos of varying compression levels with a single model; 2) effectively assist various downstream tasks on compressed videos in a plug-and-play manner; and  3) given the practical scenarios where real-time processing is required, it should meet the above objectives without causing a computation bottleneck. To achieve this, we introduce a codec-aware enhancement framework (as shown in Fig.~\ref{fig:teaser})  that reuses codec information embedded in the bitstream.  By incorporating compression factors, the framework dynamically adjusts its parameters to flexibly enhance inputs of different compression levels.  By reusing motion vectors and partition maps, it efficiently aggregates temporal and spatial clues without introducing redundant computations.
 
Specifically, our codec-aware enhancement framework comprises a compression-aware adaptation (CAA) network and a bitstream-aware enhancement (BAE) network. The CAA network serves as a ``meta'' network that dynamically adjusts the parameters of the subsequent BAE network.  A hierarchical adaptation mechanism is proposed to first estimate sequence-adaptive parameters based on the sequence compression level, and then re-weight these parameters according to the frame compression level, thereby achieving a BAE network tailored for each input frame. The frame-adaptive BAE network conducts motion vector alignment to aggregate intra-frame information and provide useful clues for the current frame.  Subsequently, based on the region complexity indicated by partition maps, the region-aware refinement assigns independent filters for different regions, achieving flexible enhancement for different regions. Comprehensive experiments are conducted to demonstrate the superiority of our method in improving the quality of compressed videos, and the effectiveness of assisting in various downstream tasks (\ie, video super-resolution, optical flow estimation, and video object segmentation). Our contributions are summarized as follows:
\begin{itemize}
\item We present a codec-aware framework for versatile compressed video enhancement, which adaptively enhances input videos of different compression levels and supports a wide range of downstream vision tasks.
\item We develop a compression-aware adaptation (CAA) network and a bitstream-aware enhancement (BAE) network that utilize the off-the-shelf codec information, contributing to generalizing across different compression settings and boosting the enhancement performance with a unified framework. 
\item Experimental results show the superiority of our method over existing enhancement methods, and its effectiveness in serving as a plug-and-play enhancement module to assist in downstream tasks. 
\end{itemize}

\section{Related Work} \vspace{-2pt}
\subsection{Compressed Video Enhancement} \vspace{-2pt}
Existing compressed video enhancement methods can be categorized into in-loop and post-processing methods.   Although in-loop methods~\cite{dai2017convolutional,pan2020efficient,huang2020frame,jia2019content} effectively improve the quality of reconstructed frames, they embed filters in the encoding and decoding loops, therefore not suitable for enhancing already compressed videos.  While the post-processing methods~\cite{8450025,sun2022quality,9591601,zhang2022dcngan, 10254565,mwgan,zhu2024cpga,jiang2023video,ma2024cvegan,ramsook2023learnt,huang2023fastcnn,10332936} provide more practical solutions to enhance compressed videos by placing filters at the decoder side.  Observing the quality fluctuation across frames, MFQE~\cite{mfqe} locates the peak quality frame (PQF) with an SVM-based detector and proposes a multi-frame quality enhancement mechanism to enhance non-PQFs.   MFQE 2.0~\cite{mfqe2} further designs a BiLSTM-based detector and performs multi-frame quality enhancement for both non-PQF and PQF. To address the inaccuracies in optical flow estimation from compressed videos, STDF~\cite{stdf} proposes estimating the offset field using spatio-temporal deformable convolution. S2SVR~\cite{lin2022unsupervised} introduces a sequence-to-sequence network to model long-range dependencies within frames.  The aforementioned methods inflexibly equip a separate model for each compression level, while we propose to adaptively handle diverse inputs with a single unified model. Recent methods~\cite{Ehrlich_2024_WACV,zhu2024cpga} utilize spatial priors from bitstream to address multiple compression levels, however, they only consider I/P-frames, whereas we design a hierarchical adaptation mechanism to address all types of frames.

\subsection{Codec-Aware Video Super-Resolution}\vspace{-2pt}
Some works in video super-resolution (VSR)~\cite{li2024enhanced,zhang2017fast,li2021comisr,chen2020bitstream,cvcp,zhang2022codec,wang2023compression} explore ways of using codec information such as motion estimation and spatial prior for reconstruction.  COMISR~\cite{li2021comisr} focuses on reducing accumulated warping errors caused by the random locations of the intra-frame from compressed video frames. Chen~\etal~\cite{chen2020bitstream} employ motion vectors to build the temporal relationship and suppress coding artifacts. CVCP~\cite{cvcp} utilizes motion vectors and spatial priors with a guided soft alignment scheme and guided SFT layer, respectively. CIAF~\cite{zhang2022codec} leverages motion vectors and residuals to model temporal relationships and skip redundant computations, respectively. Despite leveraging codec information, these methods focus on a single task (\ie, VSR). In contrast, our method not only shows competitive performance in VSR (see supplementary materials), but also effectively supports a range of downstream tasks,  which is not explored by the aforementioned works.

\subsection{Dynamic Neural Networks}\vspace{-2pt}
Instead of setting separate models for different inputs, the early mixture of expert (MoE) structure~\cite{eigen2013learning,aljundi2017expert,ma2018modeling} constructs parallel network branches and selectively executes branches to obtain the weighted outputs.   Instead of increasing the number of parallel branches, the dynamic parameters ensemble strategy~\cite{survey_dynamic,dynamicconv,yang2019condconv,eigen2013learning,aljundi2017expert,ma2018modeling} presets parallel expert layers and selectively fusing their parameters to promote the network capability and generalization ability, therefore serving as an efficient alternative to MoE.  To promote the generalization ability of pre-trained models, Gaintune~\cite{gaintuning} proposes to predict a single multiplicative scaling parameter for each channel according to test samples, thus modifying static models to test-adaptive ones. Li~\etal~\cite{se} handle the conflicts between the domain-agnostic model and multiple target domains with dynamic transfer, which is simply modeled by combining residual matrices and a static convolution matrix.  
DRConv~\cite{drconv} divides the input image into different regions with a learnable mask and assigns multiple filters for these regions, which enhances the feature representation ability without introducing a noticeable computation burden. Instead of searching and learning conditions for dynamic parameters, we leverage priors embedded in the bitstream as the condition.

 \begin{figure}[t]
\hspace{-2pt}\includegraphics[width=1\linewidth, clip=true, trim=10pt 0 14pt 32pt]{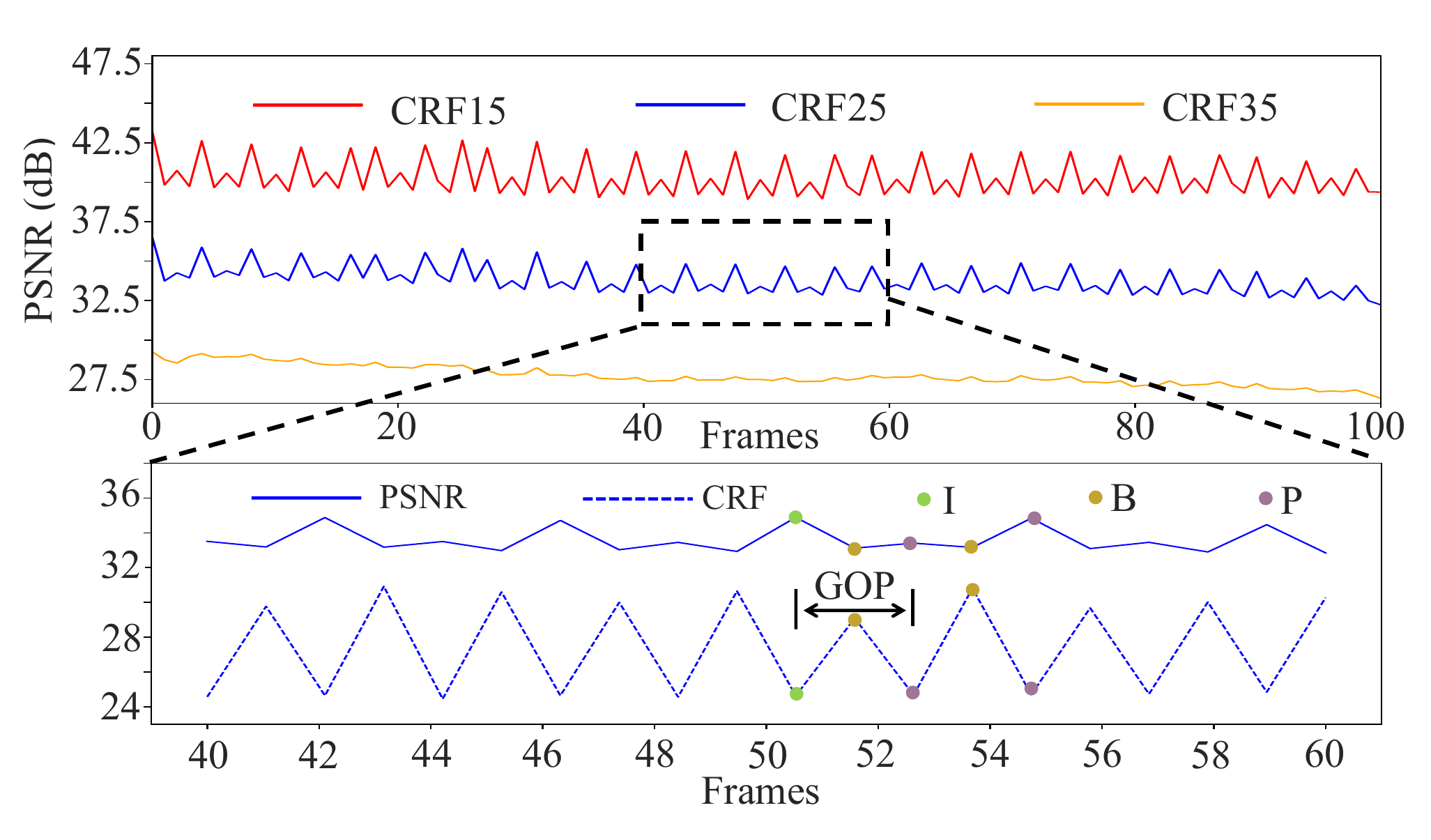}\vspace{-10pt}
    \caption{Hierarchical structure of quality adjustment, where frames are divided into multiple groups of pictures (GOP). The Constant Rate Factor (CRF) affects video quality at both sequence and frame levels. An increase in the CRF value indicates a reduction in video quality (\eg, lower PSNR values).} \vspace{-0.11in}
\label{fig:preliminarya}
\end{figure}

\vspace{-4pt}\section{Preliminaries}\label{sec:background}\vspace{-2pt}
We take H.264~\cite{h264} as a representative standard to analyze available codec information. Note recent codecs~\cite{h265,h266} also provide similar priors (\ie, CRF, motion vector, partition map), therefore assuring the applicability of our method. To reduce transmission bandwidth, codecs compress videos by adjusting quality and reducing redundancy.

\vspace{-2pt}\subsection{Hierarchical Quality Adjustment}\label{sequence and frame compression}\vspace{-2pt}
 Video quality is commonly influenced by the constant rate factor (CRF), which involves hierarchical adjustment for both sequence-wise and frame-wise compression.

\noindent \textbf{Sequence-wise CRF.}
The CRF ranges from 0 to 51 to balance compression efficiency and visual quality. A higher CRF results in more compact output but increased pixel loss  (\eg, the average PSNR of CRF35 is much lower than CRF15 in Fig.~\ref{fig:preliminarya}). By considering the sequence-wise CRF (denoted as $CRF_s$), the enhancement network can be tailored to handle videos of different compression levels.

\noindent \textbf{Frame-wise CRF  }
As shown in Fig.~\ref{fig:preliminarya}, a video sequence is divided into multiple groups of pictures (GOP) and further categorized as I-frames, P-frames, or B-frames.  The CRF value of each frame (denoted as $CRF_i$) is dynamically adjusted based on $CRF_s$ so that lower $CRF_i$ is assigned for I/P frames to maintain quality and higher $CRF_i$ for B frames for compact representations.

Inspired by the hierarchical quality adjustment paradigm, we design a hierarchical adaptation paradigm that first performs sequence adaptation to predict network parameters based on $CRF_s$, and then re-weights these parameters according to $CRF_i$ for frame adaptation.  In practical scenarios where $CRF_i$ is unavailable (\eg, limited access to full bitstream),  the proposed method can instead use slice type (I/P/B) for frame adaptation, which is demonstrated to yield similar performance in Sec.~\ref{sec:results}.

 \begin{figure*}[t]
\centering
 \vspace{-4pt}\includegraphics[width=0.88\linewidth, clip=true, trim=2pt 14pt 18pt 0]{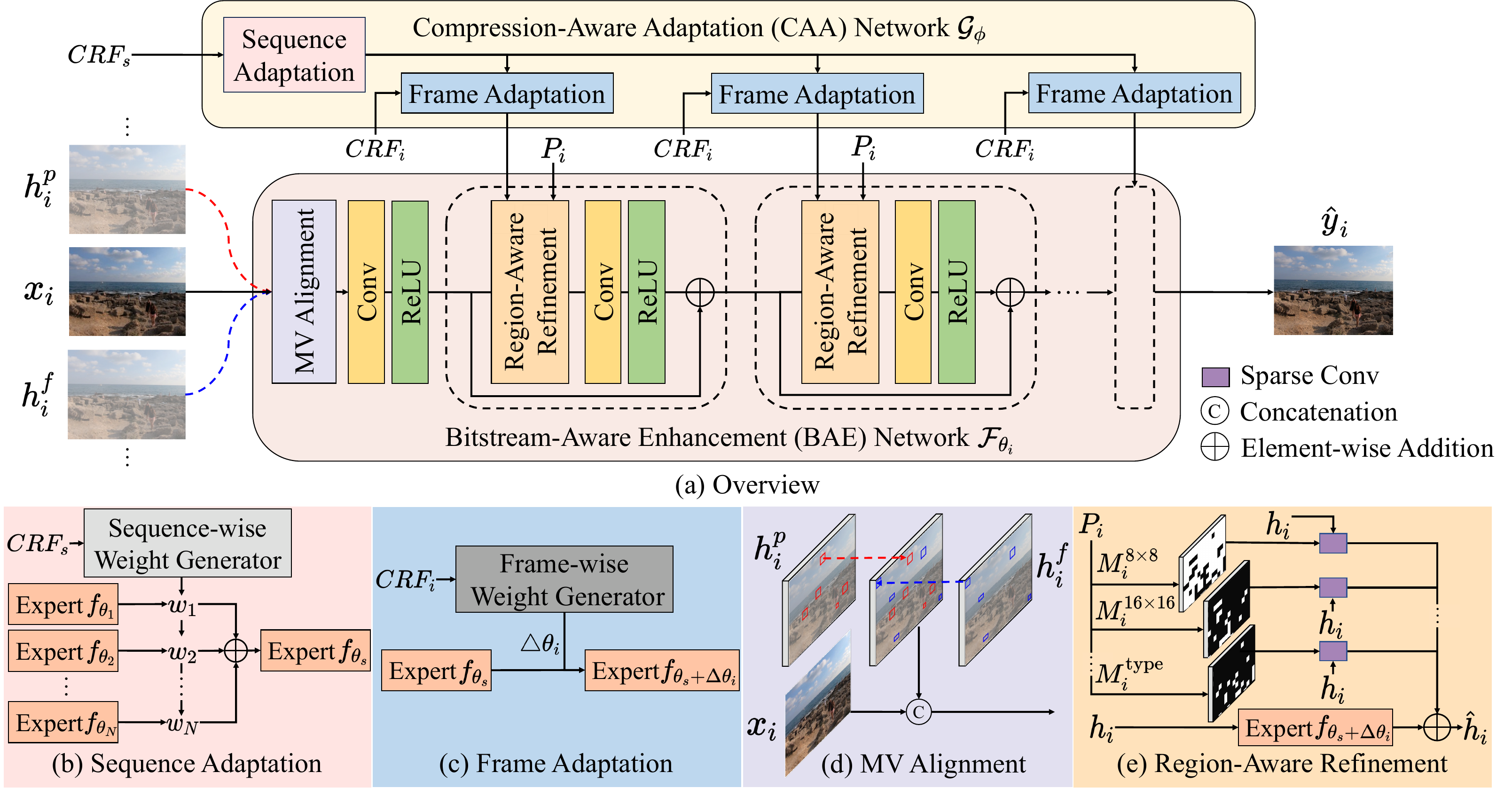} \vspace{-4pt}
\caption{The proposed Codec-Aware Enhancement Framework consists of two sub-networks: 1) the Compression-Aware Adaptation (CAA) Network, which hierarchically applies sequence adaptation and frame adaptation to dynamically adjust parameters of the enhancement network; and 2) Bitstream-Aware Enhancement (BAE) Network, which leverages motion vectors to align frames and conducts region-aware refinement to flexibly enhance regions of different complexity.  }\vspace{-0.13in}
\label{fig:overview}
\end{figure*}

\vspace{-2pt}\subsection{Redundancy Reduction}\label{sec:reduce_redun}\vspace{-2pt}
To reduce redundancy and improve the entropy of the bitstream, codecs block-wisely perform motion estimation to model the intra-frame correlations and embed the correlations in the bitstream for decoding.

\noindent \textbf{Partition map. }
As shown in Fig.~\ref{fig:teaser}, different regions of each frame are partitioned into blocks of varying sizes (\eg, H.264 provides macroblocks of 16$\times$16, 16$\times$8, 8$\times$16, and 8$\times$8) according to the texture complexity. Flat regions (\eg, the ground) can tolerate higher quantization errors and are therefore divided into blocks of large size, while complex regions (\eg, leaves and fence) take smaller blocks to maintain details. To effectively enhance regions of different complexity, we propose dynamically assigning filters based on the partition map that indicates region complexity.
  
\noindent \textbf{Motion vector. }
Motion vectors are utilized in decoding to aggregate information from reference frames and propagate information of current frame. As illustrated in Fig.~\ref{fig:teaser}, they describe the relationship between current frame and its reference frames in a block-wise manner. Although motion vectors can be noisy and are less precise than optical flow, they effectively align reference frames with current frame,  serving as a cost-effective alternative for optical flow.

\vspace{-2pt}\section{Codec-Aware Enhancement Framework} \vspace{-2pt}
\subsection{Overview}\vspace{-2pt}
As shown in Fig.~\ref{fig:overview}(a), the proposed method comprises a compression-aware adaption (CAA) network $\mathcal{G}_{\phi}$ and a bitstream-aware enhancement (BAE) network $\mathcal{F}_{\theta_i}$. The CAA network employs a hierarchical compression adaptation mechanism to estimate parameters for the frame-adaptive BAE network, which then aggregates intra-frame information and performs region-aware refinement to enhance the input compressed frame.

\vspace{-2pt}\subsection{Compression-Aware Adaptation Network}\vspace{-2pt}\label{sec:CAA}
To handle sequences of varying compression levels and quality fluctuations across frames, as illustrated in Fig.~\ref{fig:overview}(a), the CAA network $\mathcal{G}_{\phi}$ utilizes $CRF_s$  for sequence adaptation to estimate sequence-adaptive parameters, and perform frame adaptation to refine these parameters based on $CRF_i$.

\begin{figure}[t]
\centering
    \vspace{-2pt}\includegraphics[width=0.75\linewidth, clip=true, trim=0 18pt 14pt 12pt]{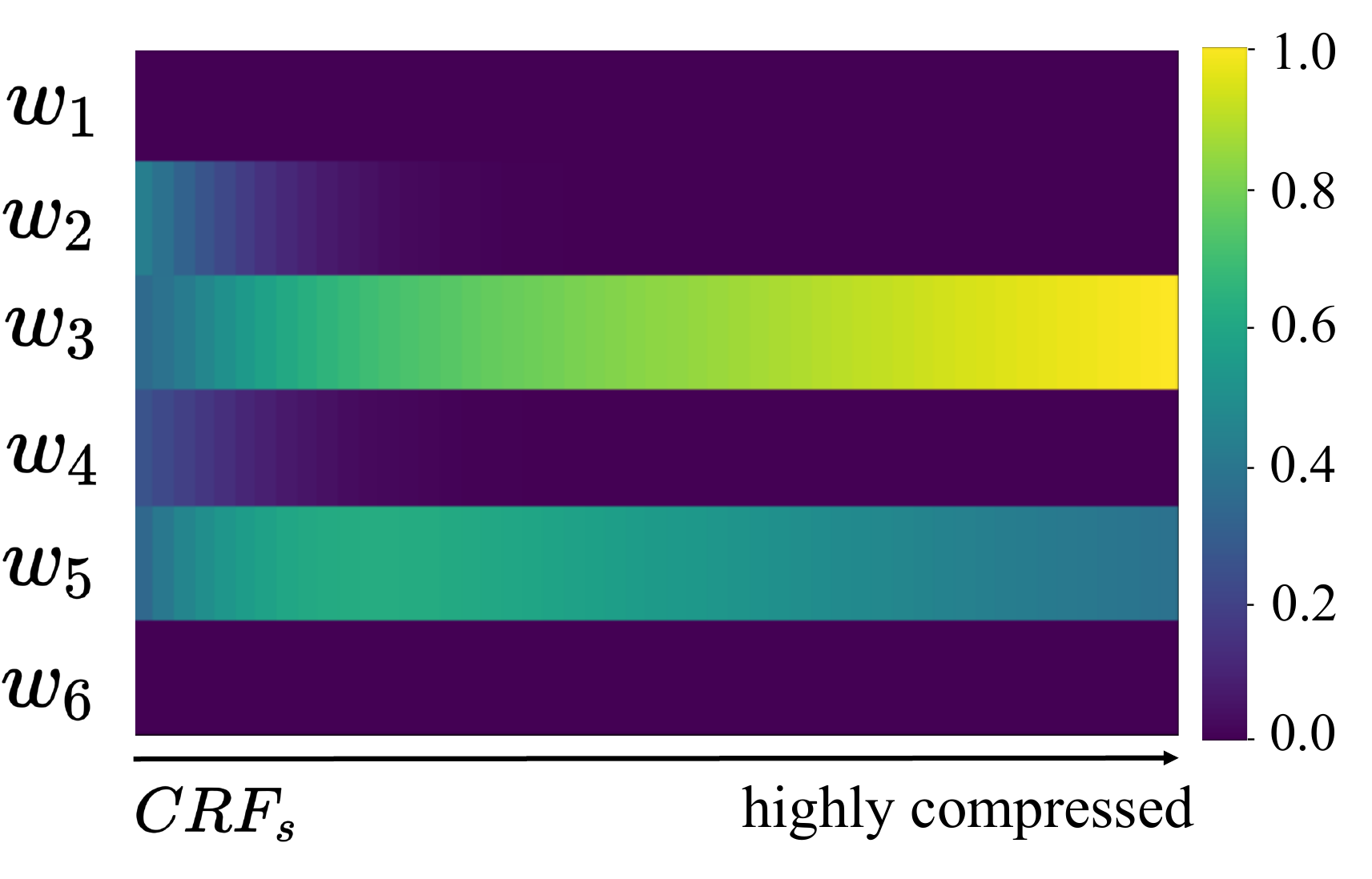}\vspace{-6pt}
\caption{Visualization of $w_n$ against different $CRF_s$, where each expert shows a distinct preference for specific $CRF_s$ values. }  \vspace{-0.1in}
\label{fig:expert vis}
\end{figure}

\noindent \textbf{Sequence adaptation. }
To ensure robust performance across multiple compression settings without increasing complexity, we propose estimating sequence-adaptive parameters for the enhancement network instead of fusing features from separate submodels.  As shown in Fig.~\ref{fig:overview}(b), parallel expert layers $\{f_{\theta_1}, f_{\theta_2}, ... , f_{\theta_N}\}$, which share the same architecture but have independent parameters, serve as the basis for parameter combination. The sequence-wise $CRF_s$ is adopted as the condition to re-weight parameters of these expert layers, which is expressed as follows, \vspace{-0.11in}
\begin{equation}\label{eq:combine2}
\hspace{-10pt}\begin{aligned}
f_{\theta_s}  & =\mathcal{G}_{\phi_s}(CRF_s,\{f_{\theta_1}, f_{\theta_2}, ...    , f_{\theta_N}\})  
 = \sum_{n=1}^N w_n f_{\theta_n},  
\end{aligned}\vspace{-0.1in}
\end{equation} 
where $f_{\theta_s}$ and $\mathcal{G}_{\phi_s}$ denote the sequence-adaptive expert layer and the sequence-wise weight generator, respectively. $w_n$ denotes the weight for each expert layer. We set $N=6$ (see ablation studies in supplementary materials) and visualize $w_n$ against different $CRF_s$ in Fig.~\ref{fig:expert vis}, which shows that each expert layer has a distinct preference for specific $CRF_s$.  Compared to MoE that re-weights output features, re-weighting expert layer parameters (as shown in Eq.~\ref{eq:combine2}) is computationally efficient and comparable to the network constructed with a single expert layer. Note that $CRF_s$ is constant for frames within the same sequence, $f_{\theta_s}$ is predicted only once and reused by subsequent frames.

\noindent \textbf{Frame adaptation. }
To flexibly enhance frames with different visual quality, we propose to re-weight the sequence-based $f_{\theta_s}$ using frame-wise $CRF_i$.  We attribute the quality fluctuation between the sequence and current frame to the disparity between $CRF_s$ and $CRF_i$, which can be addressed by introducing a set of frame-wise auxiliary parameters $\triangle{\theta_i}$.  As shown in Fig.~\ref{fig:overview}(c), the auxiliary parameters $\triangle{\theta_i}$ that conditioned on $CRF_i$ re-weights the sequence-adaptive $f_{\theta_s}$ to obtain the frame-adaptive expert layer $f_{\theta_i}$, which is expressed as follows,\vspace{-5pt}
\begin{equation}\label{eq:combine3}
\begin{aligned}
f_{\theta_i}  & = \mathcal{G}_{\phi_i}(CRF_i , f_{\theta_s}) 
 = f_{\theta_s+\triangle{\theta_i}},
\end{aligned} \vspace{-5pt}
\end{equation}
where $f_{\theta_i}$ and $\mathcal{G}_{\phi_i}$ denote the estimated frame-adaptive expert layer and the frame-wise parameters generator, respectively.  As shown by the black dashed lines in Fig.~\ref{fig:overview}(a), the obtained $f_{\theta_i}$ is used to construct the enhancement blocks, resulting in the frame-adaptive BAE network $\mathcal{F}_{\theta_i}$  (introduced in the following Sec.~\ref{sec:BAE}).

\begin{figure}[t]
\centering  
\includegraphics[width=0.86\linewidth, clip=true, trim=0 4pt 0 0]{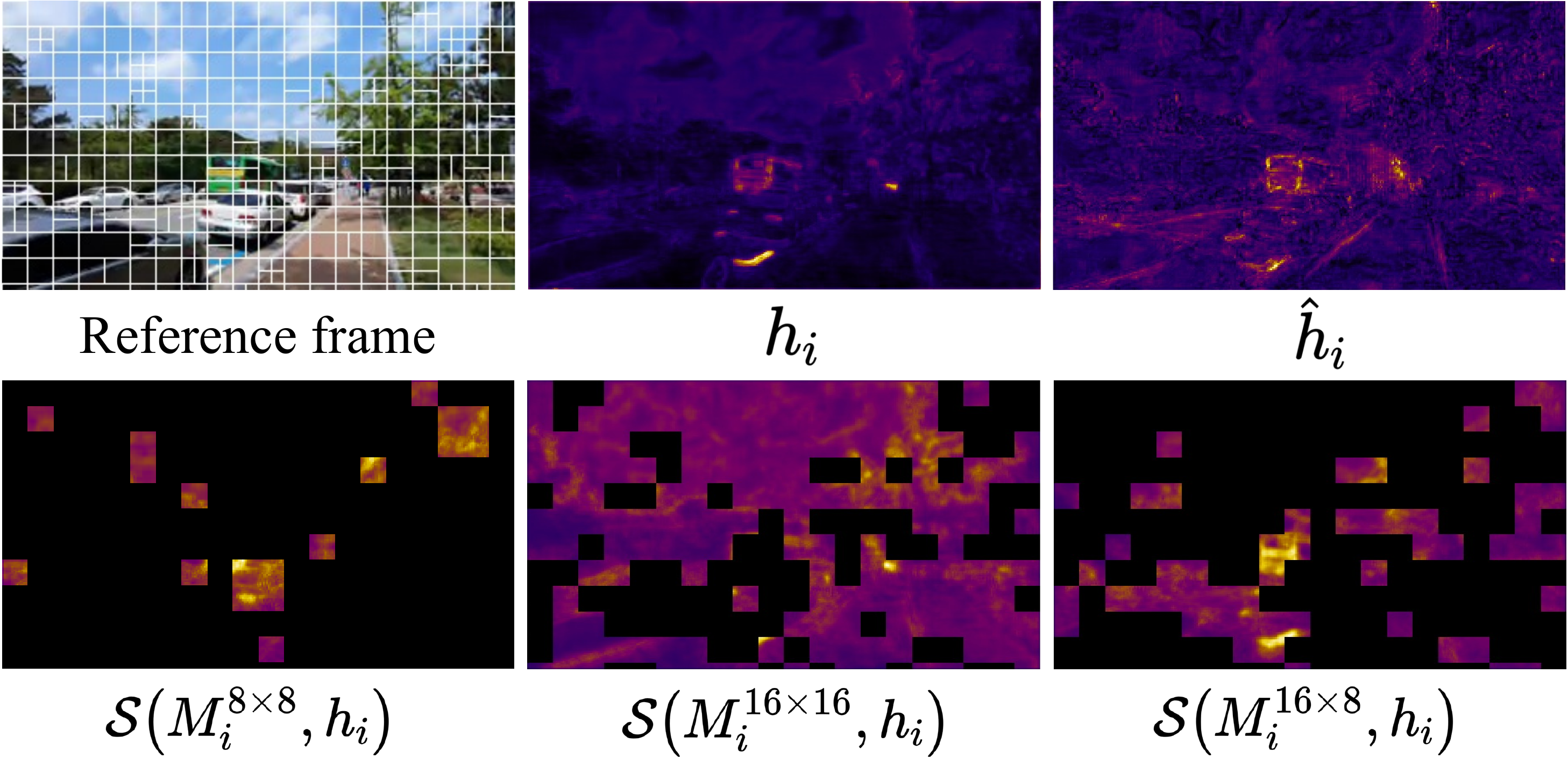}\vspace{-5pt}
\caption{Visualization of features in region-aware refinement, where $h_i$ and $\hat{h}_i$ indicate the input and output features, respectively. The refined features are denoted in the format of $\mathcal{S}(M_i^{type},  h_i)$.}\vspace{-0.13in}
\label{fig:kernel}
\end{figure} 

\subsection{Bitstream-Aware Enhancement Network}\label{sec:BAE} 
To leverage high-quality frames and propagate information, the BAE network $\mathcal{F}_{\theta_i}$ utilizes motion vectors to align reference frames with the current frame.   Meanwhile, the partition map serves as spatial complexity guidance to enable flexible enhancement of different regions.

\noindent \textbf{Motion vector alignment.}
Since the motion vectors roughly model the temporal relationship in a block-wise manner,  for each block of the current frame shown in Fig.~\ref{fig:overview}(d), motion vectors locate blocks with similar content in the previous and future reference features (highlighted with red and blue boxes). The warped reference features are concatenated with current frame along the channel dimension as input of the BAE network, expressed as follows, 
 \vspace{-8pt}\begin{equation}
\hat{x_i} = [MV(h_i^p), MV(h_i^f), x_i], \vspace{-3pt}
\end{equation}  
where $\hat{x_i}$, $h_i^p$ and $h_i^f$ denote the current input frame, enhanced features of previous and future reference frames, respectively. $MV$ denotes warping reference features based on motion vectors. $[, ]$ denotes concatenation along channel dimension.  Bilinear interpolation is adopted for the case that the offsets of motion vectors are not integers.

\noindent \textbf{Region-aware refinement. }
 To effectively enhance regions of different complexity, we propose to dynamically assign different filters for regions based on the partition map. As shown in Fig.~\ref{fig:overview}(e),  the block-based partition map $P_i$ is decoupled into multiple binary masks $\{M^{8\times8}_i, M^{16\times16}_i, ..., M^{type}_i\}$ according to the size of macroblocks, allowing separate refinement of regions using sparse convolution~\cite{liu2015sparse}. The output is defined as the sum of frame-adaptive extracted features and separately  region-aware refined features, depicted as follows,
 \vspace{-10pt}\begin{equation}\label{eq:refine}
\hat{h_i}  = \mathcal{F}_{\theta_i}(\hat{x}_i, P_i)    
  = f_{\theta_i} * h_i + \sum_{type=1}^M \mathcal{S}(M_i^{type},  h_i),  \vspace{-10pt}
\end{equation}  
where $\hat{h_i}$ indicates the output features. $\mathcal{S}$ denotes the operations applying sparse convolution guided by mask $M_i^{type}$ to refine input features. In H.264 standard, three types of macroblocks are used ($16\times16$, $8\times16/16\times8$, and $8\times8$), thus $M$ is set to 3. Features in region-aware refinement are visualized in Fig~\ref{fig:kernel}, where the refined features are denoted with operations like $\mathcal{S}(M_i^{8\times 8},  h_i)$.  As can be seen, the output features $\hat{h}_i$ contain more fine-grained and high-frequency details than $h_i$.  Meanwhile, refined features provide distinct activations for different regions.   For instance,   $\mathcal{S}(M_i^{8\times 8},  h_i)$ focuses on static objects (\eg, trees) while $\mathcal{S}(M_i^{8\times 16},  h_i)$ focuses on moving objects (\eg, the bus).

\vspace{-2pt}\subsection{Loss function}\vspace{-2pt}
We adopt Charbonnier penalty loss~\cite{charbonnierloss} as the loss function and train the proposed codec-aware enhancement framework in an end-to-end manner. The specific loss function is expressed as follows,\vspace{-8pt}
\begin{equation}
\mathcal{L}=\frac{1}{T} \sum_{i=1}^T \sqrt{\left\|y_i-\hat{y}_i\right\|^2+\epsilon^2},\vspace{-4pt}
\end{equation}
where $y_i$, $\hat{y}_i$ and $T$ indicate the uncompressed ground truth, the predicted output, and the length of the input sequence. $\epsilon$ is set to $1 \times 10^{-12}$.

\vspace{-6pt}\section{Experiments}\vspace{-2pt}
\subsection{Experimental Settings}\vspace{-2pt}
\noindent \textbf{Compression settings.}
H.264 is a popular video compression standard that compresses nearly 85\%  of internet videos~\cite{bitmovin}, and tends to introduce more severe degradations than H.265 and H.266. We adopt H.264~\cite{h264avc}  and compress videos with the $CRF_s$ values of 15, 25, and 35.

\noindent \textbf{Tasks and training dataset. } Our tasks involve quality enhancement and assisting downstream tasks on compressed inputs.   The primary downstream tasks include video super-resolution, optical flow estimation and video object segmentation, with video inpainting reported as an extension to fully evaluate the versatility.  Training splits of REDS~\cite{nah2019ntire} and DAVIS~\cite{pont20172017} datasets are combined for training.

\noindent \textbf{Compared methods.}
We compare with representative methods in compressed video enhancement, including MFQE 2.0~\cite{mfqe2}, STDF~\cite{stdf}, S2SVR~\cite{lin2022unsupervised} and Metabit~\cite{Ehrlich_2024_WACV}. For a fair comparison, we fully retrain these methods with the same training dataset and configurations.   For downstream tasks, the compressed video is first enhanced by quality enhancement methods, and then fed to downstream models for corresponding tasks and further assessment.

  \begin{table*}[t]
\centering
\Large
 \vspace{-6pt}\resizebox{1\linewidth}{!}{
\begin{tabular}{l|c|c|c|c|c|c|c|c|c|c}
\toprule
\multirow{2}{*}{Method} &  \multirow{2}{*}{Param/M}  &  \multirow{2}{*}{FLOPs/G}  &   \multirow{2}{*}{Speed/ms}  & \multirow{2}{*}{FPS} & CRF15& CRF25 & CRF35  & CRF18   & CRF28 & CRF38   \\
\cmidrule{6-11}
& & & & &   PSNR$\uparrow$  /  SSIM$\uparrow$   &  PSNR$\uparrow$  /  SSIM$\uparrow$ &  PSNR$\uparrow$  /  SSIM$\uparrow$  & PSNR$\uparrow$  /  SSIM$\uparrow$   &  PSNR$\uparrow$  /  SSIM$\uparrow$ &  PSNR$\uparrow$  /  SSIM$\uparrow$  \\
\midrule  
Input &  - &  - &  - & - &  41.04 / 0.9785 & 34.92 /  0.9363 & 29.25 / 0.8238 &   39.12 / 0.9698  &  33.18 / 0.9123	 & 27.69 / 0.7725 \\ 
 MFQE 2.0~\cite{mfqe2} & 1.64 	& 51 &	53 & 	19 &  40.95 / 0.9806 & 34.83 / 0.9378	 & 29.22 / 0.8256 & 38.97 / 0.9712 & 	 33.13 / 0.9140 &  27.67 / 0.7742  \\
 STDF~\cite{stdf}  &   1.27  & 45	& 38	& 26   &  41.15 /	0.9793  &   35.23 /	0.9398	 & 29.74 / 0.8359 &  39.28 / 0.9712 & 33.58 / 0.9178 	&  28.11 / 0.7853  \\ 
  S2SVR~\cite{lin2022unsupervised} &   7.43 	&  294 	 & - & 	-    & 41.96 / 0.9834	 &  35.61 / 0.9445 & 29.87 / 0.8391 & 39.88 / 0.9755 &  33.87 / 0.9223 & 28.19 / 0.7881\\
     Metabit~\cite{Ehrlich_2024_WACV} &   1.60 &  92  & 24	 & 42   &   41.04  /	0.9785  & 34.92 / 0.9363  & 29.25	/ 0.8238  &  39.11 / 0.9698	& 33.18 / 0.9123 &	27.69	/	0.7725 \\ 
 \midrule
\multirow{2}{*}{Ours} &  \multirow{2}{*}{4.56} &  \multirow{2}{*}{47}	&  \multirow{2}{*}{36} & \multirow{2}{*}{28} 
 & \cellcolor{lightgrey!40}\underline{42.22} / \underline{0.9842}  &\cellcolor{lightgrey!40}\underline{35.90} / \underline{0.9468}  &\cellcolor{lightgrey!40}\underline{30.17} / \underline{0.8471} &  \cellcolor{lightgrey!40}\underline{40.17} / \underline{0.9767} &\cellcolor{lightgrey!40}\underline{34.16} / \underline{0.9258} & \cellcolor{lightgrey!40}\underline{28.49} / \underline{0.7985} \\
 &  &  &  & & \textbf{42.24 / 0.9842}	 & \textbf{35.91 / 0.9468}	 & \textbf{30.19 / 0.8472}	& \textbf{40.18 / 0.9767}	 & \textbf{34.17 / 0.9258}	 & \textbf{28.52 / 0.7985}  \\
\bottomrule
\end{tabular}
}\vspace{-6pt}
\caption{Quantitative results on quality enhancement, where PSNR and SSIM (higher is better) are adopted for evaluation.  The best and second best results are marked with \textbf{bold} and \underline{underline}. Results obtained by replacing $CRF_i$ with slice type are highlighted with \colorbox{lightgrey!40}{grey}. } \vspace{-0.13in}
\label{tab:QE}
\end{table*}

\subsection{Results}\vspace{-2pt}\label{sec:results}
 Our evaluations are two-fold: 1) verifying the quality enhancement performance on seen, unseen, and highly compressed scenarios; 2) evaluating the versatility to assist different downstream tasks on multiple compression settings.

  \begin{figure*}[t]
\centering
\includegraphics[width=1\linewidth, clip=true, trim=0 10pt 0 0]{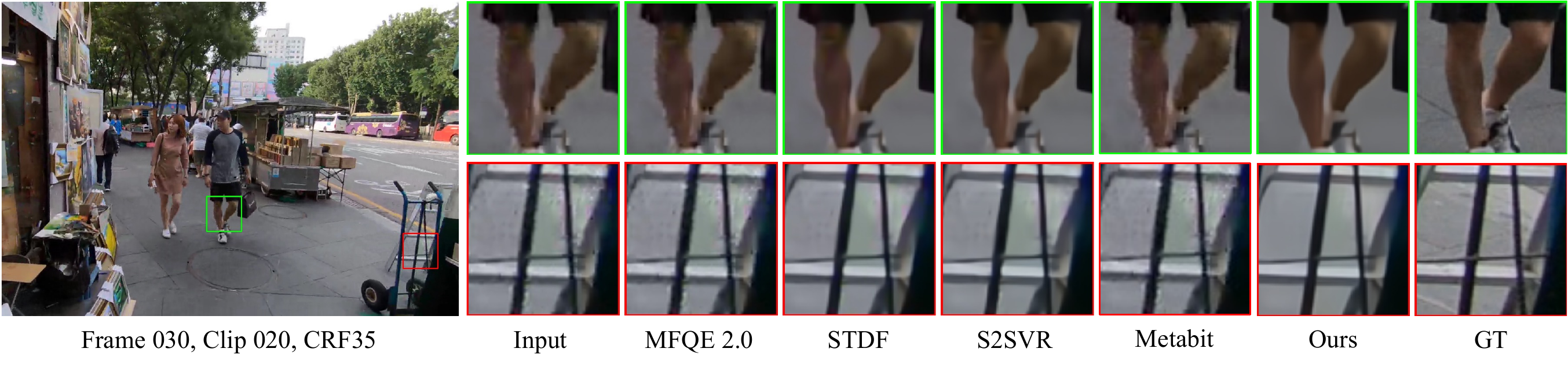}\vspace{-6pt}
\caption{Qualitative results on quality enhancement, where our method effectively reduces the compression artifacts, achieving visually pleasant results. In contrast, the results of the compared methods still contain severe distortions (\eg, the calf in the 1st row).}
\label{fig:enhance}\vspace{-0.13in}
\end{figure*}

\begin{table}[t]
\centering
\Large
\resizebox{0.9\linewidth}{!}{
\begin{tabular}{l|c|c|c}
\toprule
\multirow{2}{*}{Method}& CRF15& CRF25 & CRF35  \\
\cmidrule{2-4}
& PSNR$\uparrow$  /  SSIM$\uparrow$  & PSNR$\uparrow$  /  SSIM$\uparrow$  &  PSNR$\uparrow$  /  SSIM$\uparrow$     \\
\midrule  
BasicVSR  & 29.24 / 0.8212	& 26.19 / 0.7131	& 23.40 / 0.6005 \\
+ MFQE 2.0  &  29.29 / 0.8233 &	26.28 / 0.7182 &	23.46 / 0.6056 \\
+ STDF &  29.31 / 0.8247 &	26.51 / 0.7293	&23.80 / 0.6249 \\
+ S2SVR & \underline{29.45} / \underline{0.8288}	&  \underline{26.70} / \underline{0.7346}	&  \underline{23.86} / \underline{0.6270}   \\
 + Metabit &29.24 /	0.8211  &	26.19 /	0.7131 &	23.39  /	0.6005  \\
+ Ours & \textbf{29.54 / 0.8328}	& \textbf{26.85 / 0.7419}	& \textbf{24.02 / 0.6361}   \\ 
\midrule 
IconVSR &  29.29 / 0.8230	& 26.19 / 0.7130	& 23.39 / 0.6003 \\
+ MFQE 2.0 &   29.37 / 0.8254	& 26.28 / 0.7182	& 23.45 / 0.6055 \\
+ STDF  &  29.36 / 0.8263 & 	26.52 / 0.7292 &	23.79 / 0.6248 \\
+ S2SVR  &   	\underline{29.54} / \underline{0.8306}	&  \underline{26.71} / \underline{0.7345}	&  \underline{23.85} / \underline{0.6269}    \\
 + Metabit & 29.29 /	0.8230	& 26.19 /	0.7130	& 23.39 /	0.6003 \\
+ Ours & \textbf{29.63 / 0.8344} &	\textbf{26.86 / 0.7418} &	\textbf{24.01 / 0.6360}  \\
\midrule 
BasicVSR++  & 29.61 / 0.8303 & 26.19 / 0.7118	 & 23.38 / 0.5998  \\
+ MFQE 2.0 & 29.66 / 0.8322	& 26.27 / 0.7169	 & 23.44 / 0.6051 \\
+ STDF  &  29.68 / 0.8338	 & 26.53 / 0.7289	 & 23.79 / 0.6247 \\
+ S2SVR &   \underline{29.82} / \underline{0.8371} & \underline{26.72} / \underline{0.7346} & \underline{23.85} / \underline{0.6269} \\
 + Metabit & 29.61 /	0.8303	& 26.19 /	0.7118	& 23.38	/ 0.5997 \\
+ Ours & \textbf{29.92 / 0.8407}	 & \textbf{26.87 / 0.7419} & 	\textbf{24.00 / 0.6358}   \\
\bottomrule
\end{tabular}
}\vspace{-6pt}
\caption{Quantitative results of $\times$4 VSR, where the best and second best results are highlighted with \textbf{bold} and \underline{underline}. }\vspace{-0.25in}
\label{tab:vsr}
\end{table}

\subsubsection{Quality Enhancement Performance}\label{Quality}
\noindent \textbf{Quantitative results.}
The results of compressed video quality enhancement are reported in Tab.\ref{tab:QE}, which is evaluated on the REDS4 datasets~\cite{nah2019ntire} using PSNR and SSIM (the higher the better).  Note the CRF values of 18, 28 and 38 are not included during training.   For each method, we include the model complexity and inference speed.  For our method, we report results of both applying $CRF_i$ and its substitution with slice type (highlighted with \colorbox{lightgrey!40}{grey}).  As shown in Tab.~\ref{tab:QE},  leveraging slice type yields comparable performance to $CRF_i$, with a negligible decrease of PSNR  ($<$0.03 dB), demonstrating the feasibility of replacing $CRF_i$ with slice type in practical.  The proposed method notably improves the quality of compressed input, achieving a PSNR gain of 1.2 dB on CRF15, while MFQE 2.0 and Metabit lead to no improvement. With similar computation cost and inference speed, our method significantly outperforms STDF, obtaining a PSNR gain of 1.09 dB on CRF15. Compared to S2SVR, our approach takes only 61\% of the parameters and 16\% of the FLOPs, and achieves a throughput of 28 FPS, which underlines its practicality.  In addition, our method shows robustness and generalization ability 
on unseen scenarios (\ie, CRF18, CRF28 and CRF38),  achieving up to 1.06 dB PSNR gain on CRF18. In contrast, other methods trained with mixed compression settings show sub-optimal performance. For instance, STDF and S2SVR only achieve PSNR gains of 0.16 dB and 0.76 dB at CRF18, while MFQE 2.0 shows no improvement. Quantitative results on highly compressed scenarios (\ie, CRF40, CRF45, CRF48)  are included in the supplementary materials.
 
\noindent \textbf{Qualitative results.}
Qualitative comparisons are provided in Fig.~\ref{fig:enhance}. As can be seen,  MFQE 2.0 and Metabit struggle to improve the quality of compressed inputs. Both STDF and S2SVR cannot adequately remove compression artifacts  (\eg, boundary of the calf), while the proposed method effectively eliminates the compression artifacts, preserving accurate edges and textures. We provide more qualitative comparisons in the supplementary materials.

\begin{figure*}[t]
\centering
 \vspace{-6pt} \includegraphics[width=1\linewidth, clip=true, trim=0 10pt 0 0]{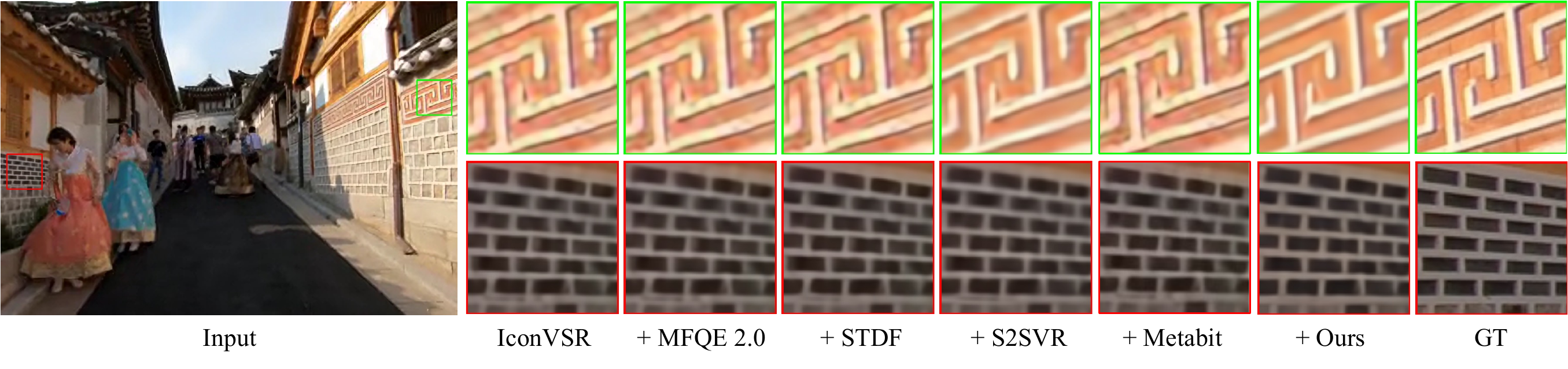}\vspace{-6pt}
\caption{Qualitative results of $\times$4 VSR.  Pre-enhancing with our method
before VSR effectively avoids amplifying compression artifacts.   While other methods cannot fully eliminate the artifacts and even severe the distortions (\eg, results of MFQE 2.0).
}   \vspace{-4pt}
\label{fig:vsr}
\end{figure*}

\vspace{-1pt}\subsubsection{Versatility Evaluation} \vspace{-1pt}
To evaluate the versatility in assisting practical downstream tasks, we employ the implementation that utilizes slice type for frame adaptation (as described in Sec.~\ref{sequence and frame compression}) to enhance compressed inputs for downstream tasks. More qualitative comparisons are included in the supplementary materials.

\noindent \textbf{Video super-resolution.} We adopt BasicVSR~\cite{chan2021basicvsr}, IconVSR~\cite{chan2021basicvsr}, and BasicVSR++~\cite{chan2022basicvsr++} as the representative baseline methods for $\times 4$ video super-resolution (VSR), which are trained on ``clean" data without considering compresstion. The evaluation is conducted on the REDS4 dataset~\cite{nah2019ntire} and summarized with PSNR and SSIM (the higher the better).  As depicted in Tab.~\ref{tab:vsr}, pre-enhancing compressed inputs with Metabit fails to improve the performance of downstream VSR models. In contrast, pre-enhancing with our framework yields consistent improvement especially in scenarios of high compression  (\eg, up to 0.62 dB PSNR gain with BaiscVSR++ on CRF35).   Meanwhile, our method significantly outperforms MFQE 2.0 and STDF, achieving PSNR gains of 0.25 dB and 0.23 dB over MFQE 2.0 and STDF on BasicVSR/CRF15, respectively. Compared with S2SVR, the proposed method offers more effective support to VSR models with lower complexity.   As shown in Fig.~\ref{fig:vsr}, performing VSR on compressed data inevitably amplifies compression artifacts (\eg, the 1st column), while results pre-enhanced with our method maintain accurate edges and textures, avoiding distortions seen in other methods.

\begin{table}[t]
\centering 
\resizebox{0.84\linewidth}{!}{
\begin{tabular}{l|c|c|c}
\toprule
\multirow{2}{*}{Method}& CRF15 & CRF25 & CRF35    \\
\cmidrule{2-4}
 &  EPE$\downarrow$  /  F1-all$\downarrow$   &  EPE$\downarrow$  /  F1-all$\downarrow$    &  EPE$\downarrow$  /  F1-all$\downarrow$    \\
\midrule
RAFT   &   5.26 / 17.81  & 7.37 / 22.13 & 16.73 / 44.70  \\
+ MFQE 2.0 & 5.32 / 17.83	 &  7.27 / 21.92	&  16.68 / 44.74\\
+ STDF  &  5.34 / 17.93 & 	 7.13 / 22.16 & 	 15.92 / \underline{44.04} \\
+ S2SVR &  \underline{5.22} / \underline{17.71}	&  \underline{6.90} / \underline{21.57}	& \underline{15.73} / 44.62\\
 + Metabit & 5.32 / 17.86	& 7.33 / 22.07	& 16.69 / 44.28 \\
+ Ours &   \textbf{5.20 / 17.69} & \textbf{6.52 / 20.99} & \textbf{14.84 / 42.56}  \\
\midrule
DEQ &   3.99 / 13.71   & 5.40 / 17.33  & 13.94 / 41.63    \\
+ MFQE 2.0  &  3.97 / 13.73	&  \underline{5.14} / 17.06 & 	14.06 / 41.72\\
+ STDF  &  4.08 / 13.84	 &  5.27 / 17.38	 &  \underline{13.52}  / \underline{40.75} \\
+ S2SVR  &  4.01 / \underline{13.69}	& 5.22 / \underline{16.99}	& 13.74 / 41.65 \\
 + Metabit  & \textbf{3.92} / 13.78	& 5.30 / 17.33	& 13.84	/ 41.21 \\
+ Ours &  \underline{3.97} / \textbf{13.68} & \textbf{4.96 / 16.54} & \textbf{13.09 / 39.43} \\
\midrule
KPAFlow & 4.46 / 16.07 & 6.71 / 20.96 & 16.50 / 45.13    \\
+ MFQE 2.0 &  \underline{4.42} / \underline{15.96}  & 	 6.71 / 20.92  & 	16.70 / 45.29 \\
+ STDF  & 4.52 / 16.19	& 6.96 / 21.53 & \underline{15.62}  /  \underline{44.17}  \\
+ S2SVR &  \textbf{4.37 / 15.96}	& \underline{6.10} / \underline{19.76}	& 15.62 / 44.23 \\
 + Metabit  &   4.47	/ 16.17	 &  6.75 / 21.08	 & 16.68 / 44.90\\
+ Ours &  4.43 / 16.10 & \textbf{5.59 / 18.73} & \textbf{14.83 / 41.24} \\
\bottomrule
\end{tabular}
}\vspace{-6pt}
\caption{Quantitative results of optical flow estimation, where we highlight the best and second best results with \textbf{bold} and \underline{underline}.} \vspace{-0.22in}
\label{tab:flow}
\end{table}

\begin{figure*}[t]
\centering
 \vspace{-4pt}\includegraphics[width=1\linewidth, clip=true, trim=0 16pt 0 0]{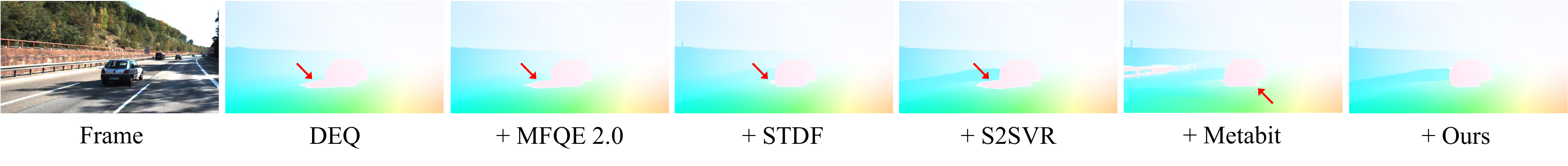}\vspace{-6pt}
\caption{Qualitative results of optical flow estimation. As can be seen, pre-processing with our method effectively corrects the mispredicted optical flow, especially the boundaries of moving objects (\eg, edge of the moving car).   } \vspace{-0.15in}
\label{fig:flow}
\end{figure*}

\begin{table}[t]
\centering
\Large
\vspace{-2pt}\resizebox{1\linewidth}{!}{
\begin{tabular}{l|c|c|c}
\toprule
\multirow{2}{*}{Method} &CRF15 & CRF25 & CRF35 \\
\cmidrule{2-4}
&  Avg$\uparrow$  /  $\mathcal{J}\uparrow$  /  $\mathcal{F}\uparrow$  &  Avg$\uparrow$   /  $\mathcal{J}\uparrow$  /  $\mathcal{F}\uparrow$  &  Avg$\uparrow$   /  $\mathcal{J}\uparrow$  /  $\mathcal{F}\uparrow$ \\
\midrule
STCN   & 85.07	   /  81.83 / 88.32    & 84.35 /  80.96 /  \underline{87.74}     & 79.20   /  76.04 / 82.37 \\ 
+ MFQE 2.0 & 84.96   /  81.71 / 88.21 &  84.30  /  80.92 / 87.69  & 79.28   /  76.11 / 82.44 \\
+ STDF  & 85.01   /  81.73 / 88.28 & 
84.23  /   80.97  / 87.50 & 
 79.77  /  76.52 / 83.02  \\
+ S2SVR &  \underline{85.17} / \underline{81.93} / \underline{88.41} & \underline{84.46} / \underline{81.20} / 87.72 & \underline{80.04} / \underline{76.88} / \underline{83.20}   \\
+ Metabit & 84.56 	/ 80.97  / 	88.14  & 	83.86  / 80.18  / 87.55  & 	79.03 / 	75.65 / 82.40  \\ 
+ Ours  &  \textbf{85.21 / 81.99 / 88.44} & \textbf{84.63 / 81.42 / 87.85} & \textbf{81.57 / 78.46 / 84.69}    \\ 
\midrule
DeAoT  & 85.90 / 82.89 / 88.91    & 85.18 / 82.37 / 88.00    & 82.87 /   \underline{79.86}  / 85.88  \\ 
+ MFQE 2.0  &85.86 / 82.84 / 88.88 &
 \underline{85.20} / \underline{82.38}  / 88.03 &
82.86 / 79.86 / 85.85 \\
+ STDF  & 85.83 / 82.80 / 88.87  
& 85.18 / 82.27 /  \underline{88.09} 
& \textbf{82.90 / 79.92 }/ \underline{85.89}  \\
+ S2SVR &   \underline{86.05} / \underline{83.09} / 89.01 & 85.05 / 82.07 / 88.04 & 82.64 / 79.63 / 85.65  \\
 + Metabit  & 85.47 / 82.04 / 88.90  & 84.95 / 	81.57 / 	\textbf{88.33}  & 82.32 / 	79.04 / 	85.59 \\
+ Ours  & \textbf{86.08 / 83.13 / 89.03} & \textbf{85.31} / \textbf{82.38} / \underline{88.25} &  \underline{82.88} / 79.83 /  \textbf{85.92} \\ 
\midrule
QDMN &85.16 / 82.20 / 88.11  &  \underline{84.16} / \underline{81.20}  / 87.12   & 79.39 / 76.61 / 82.18  \\ 
+ MFQE 2.0 & 85.13 / 82.20 / 88.06 &  84.15 / 81.18 /  \underline{87.13}  & 79.51 / \underline{76.75} / 82.27 \\
+ STDF  & \underline{85.32} / \underline{82.38} / \underline{88.27} &  83.36 / 80.44 / 86.27 & \underline{79.64} / 76.69 / \underline{82.59} \\
+ S2SVR & 85.28 / 82.32 / 88.23 & 83.64 / 80.65 / 86.63 & 79.02 / 76.15 / 81.89 \\
 + Metabit & 84.50 / 	81.14 / 	87.87  & 83.68 / 	80.30 / 	87.06 & 79.47 / 	76.41 / 	82.52 \\
+ Ours  &   \textbf{85.34 / 82.41 / 88.27} & \textbf{84.37 / 81.42 / 87.32} &  \textbf{79.78 / 76.92 / 82.65}  \\ 
\bottomrule
    \end{tabular}
}\vspace{-6pt}
\caption{Quantitative results of VOS, where the best and second best results are highlighted with \textbf{bold} and \underline{underline}. } \vspace{-0.2in}
\label{tab:vos}
\end{table}

\noindent \textbf{Optical flow estimation. }  We adopt RAFT~\cite{teed2020raft}, DEQ~\cite{deq-flow}, and KPAFlow~\cite{kpa} as baseline models for optical flow estimation.    Evaluation on the KITTI-2015 dataset~\cite{geiger2013vision} is summarized with EPE (end-point-error) and F1-all loss, where lower values indicate better accuracy. As shown in Tab.~\ref{tab:flow}, our method consistently reduces the EPE and F1-all loss across all baseline models, demonstrating its effectiveness in improving optical flow estimation.  In contrast, methods such as MFQE 2.0, STDF and Metabit fail to deliver consistent improvements. For instance,  MFQE 2.0 fails to improve the performance of RAFT on CRF15, STDF and Metabit detrimentally affects the performance of DEQ and KPAFlow on CRF15. Visualizations of predicted optical flow are shown in Fig.~\ref{fig:flow}, where inaccurate boundaries are highlighted with red arrows.  As can be seen, optical flow estimated from compressed inputs contains inaccurate boundaries, especially near-motion ones.  The proposed method helps to deliver more accurate results in these regions compared to others. For instance, it effectively corrects the optical flow near the car that was mispredicted by DEQ, while MFQE 2.0 and S2SVR provide limited improvement.

\begin{figure*}[t]
\centering
 \vspace{-6pt} \includegraphics[width=1\linewidth, clip=true, trim=0 18pt 0 0]{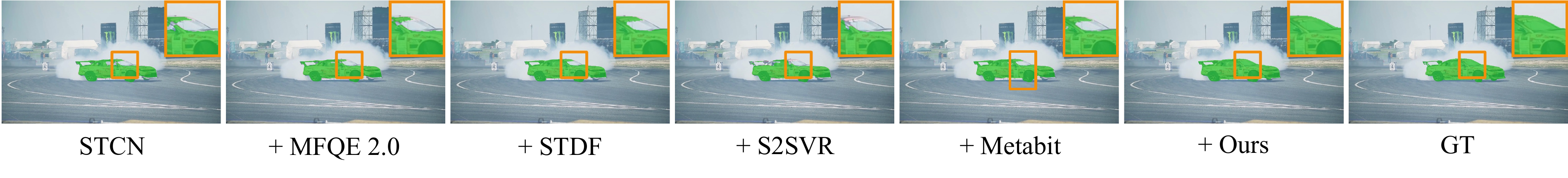}\vspace{-6pt}
\caption{Qualitative results of VOS. As can be seen, directly performing VOS on compressed inputs leads to inaccurate masks, whereas pre-enhancing with our method effectively improves the accuracy, especially for the regions of irregular shapes (\eg, the windshield).}\vspace{-0.05in}
\label{fig:vos}  
\end{figure*}

  \begin{figure*}[t]
    \centering  
   \includegraphics[width=1\linewidth, clip=true, trim=0 18pt 0 0]{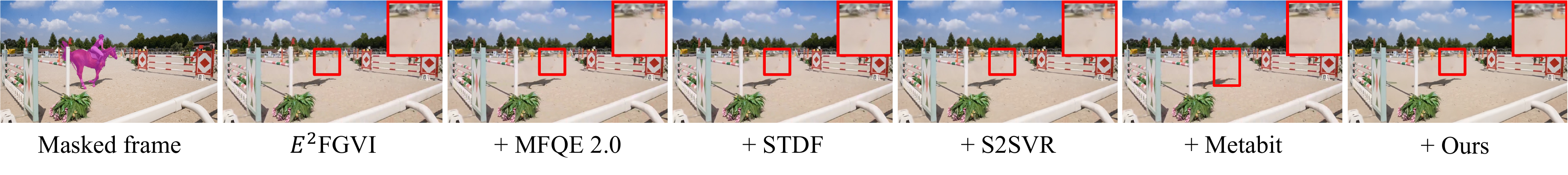}\vspace{-6pt}
    \caption{Qualitative results of video inpainting. As can be seen, pre-enhancing the compressed inputs with the proposed method helps to reduce the artifacts and color distortion in the removed region, providing more visually pleasant results. } \vspace{-0.1in}
    \label{fig:inpainting}
\end{figure*}

\noindent\textbf{Video object segmentation.} For video object segmentation (VOS), we adopt STCN~\cite{cheng2021stcn}, DeAoT~\cite{yang2022deaot} and QDMN~\cite{qdmn} as representative baselines.   Evaluations on DAVIS-17 val dataset~\cite{pont20172017} are summarized with the following metrics: the $\mathcal{J}$ (average IoU), the $\mathcal{F}$ score (boundary similarity), and the average of the above metrics (denoted as $Avg$). Higher values indicate better segmentation accuracy.  As shown in Tab.~\ref{tab:vos},  the proposed method shows the best performance in improving accuracy across VOS models.  For instance, it elevates the average accuracy for up to 2.37\% (79.20\% to 81.57\%) on STCN at CRF35, while MFQE 2.0, STDF and S2SVR yield limited improvement of 0.08\%, 0.57\% and 0.84\%, respectively. And Metabit provides no improvement on STCN at CRF35. The results of VOS are included in Fig.~\ref{fig:vos}, where accurately segmenting objects in compressed videos is challenging for VOS baselines (\eg, inaccurate mask of the windshield predicted by STCN).  Pre-enhancing the compressed videos with  MFQE 2.0 and S2SVR struggles to address this issue,  whereas the proposed method significantly refines the segmentation results, demonstrating its effectiveness in assisting the VOS task.

\noindent\textbf{Video inpainting.}  We take E$^2$FGVI~\cite{li2022towards} as the video inpainting model and perform video object removal on DAVIS-17 val dataset~\cite{pont20172017}.  The qualitative results are shown in Fig.~\ref{fig:inpainting}.    As can be seen, compression-included misalignment between objects and masks hinders the ability to remove specified objects, causing color distortions (\eg, the horse region). Pre-enhancing the compressed inputs with the proposed method notably refines the artifacts and distortions, yielding more visually pleasing results.

\subsection{Ablation Studies}\label{sec:ablation}
We start with a baseline that concatenates reference frames and current frames as input, without using codec information. We then progressively equip the baseline with  MV alignment, region-aware refinement, sequence adaptation, and frame adaptation to assess their contributions.

\begin{table}[t]
\centering
\Large
\resizebox{1\linewidth}{!}{
\begin{tabular}{l|c|c|c}
\toprule
\multirow{2}{*}{Model} &  \text { CRF15 } & \text { CRF25 } & \text { CRF35 } \\ 
\cmidrule{2-4}
& PSNR$\uparrow$ / SSIM$\uparrow$  & PSNR$\uparrow$ / SSIM$\uparrow$  & PSNR$\uparrow$ / SSIM$\uparrow$  \\
\midrule
Baseline & 41.04 / 0.9785 &	34.92 / 0.9363	& 29.25 / 0.8238 \\
\midrule
+ MV Align.& 41.69 / 0.9821	& 35.59 / 0.9437	& 29.95 / 0.8403 \\
+ RA Refine. & 42.04 / 0.9837	& 35.70 / 0.9449	& 30.00 / 0.8427 \\
\midrule
+ Seq. Adapt. & 42.08 / 0.9838 & 35.76 / 0.9458 & 30.04 / 0.8444\\
+ Frame Adapt. &  42.14 / 0.9839	 & 35.81 / 0.9460 & 	30.09 / 0.8446\\
\bottomrule
\end{tabular}
} \vspace{-6pt}
\caption{Ablation studies on MV alignment, region-aware refinement, sequence adaptation, and frame adaptation.
}\vspace{-0.13in}
\label{tab:BAE ablation}
\end{table}

\noindent \textbf{MV alignment.} As shown in the 2nd row of Tab.~\ref{tab:BAE ablation}, incorporating MV alignment yields a PSNR gain of up to 0.65 dB on CRF15, demonstrating the effectiveness of MV in aligning reference frames and current frame.

\noindent \textbf{Region-aware refinement. }
We further incorporate the region-aware refinement module to refine the features of different regions. As shown in the 3rd row of Tab.~\ref{tab:BAE ablation}, it leads to notable PSNR gains of 0.35dB, 0.11dB and 0.05dB on CRF15, CRF25 and CRF35, respectively.

\noindent \textbf{Sequence adaptation. }
 As shown in the 4th row of Tab.~\ref{tab:BAE ablation}, sequence adaptation brings PSNR gains of 0.04 dB, 0.06 dB and 0.04 dB on CRF15, CRF25 and CRF35, respectively.

\noindent \textbf{Frame adaptation.}
As shown in the 5th row of Tab.~\ref{tab:BAE ablation}, frame adaptation improves PSNR by 0.06 dB 0.05 dB and 0.05 dB on CRF15, CRF25 and CRF35, respectively.   We further analyze its effectivness on improving the temporal consistency in the supplementary materials.

\section{Conclusion}\vspace{-4pt}
In this paper, we introduce a versatile codec-aware enhancement framework that adaptively handles diverse compression settings and serves as a plug-and-play enhancement module to consistently boost various downstream tasks. By reusing the off-the-shelf codec information, our method minimizes additional computational costs. Compared with existing compressed video enhancement solutions, it shows superority in both enhancement performance and robustness, making it possible to deploy pre-trained models on compressed videos without a significant performance drop. 

\noindent\textbf{Acknowledgments.} We acknowledge funding from the National Natural Science Foundation of China under Grants 62131003 and 62021001.
 
{
    \small
    \bibliographystyle{ieeenat_fullname}
    \bibliography{egbib}
}

\appendix
\clearpage
\setcounter{page}{1}
\maketitlesupplementary

\noindent This supplementary document is organized as follows:

– Section~\ref{sec:procedure} provides a detailed explanation and pseudo-code to clarify the procedure for enhancing compressed frames.

– Section~\ref{sec:Generalization} reports quantitative comparisons for quality enhancement in highly compressed scenarios (\ie, CRF40, CRF45 and CRF48) to demonstrate the robustness of the proposed method.

– Section~\ref{sec:Qualitative Comparisons} provides more qualitative comparisons on quality enhancement (Section~\ref{sec:Quality Enhancement}) and downstream tasks (Section~\ref{sec:Versatility Evaluation}), including video super-resolution, optical flow estimation, video object segmentation, and video inpainting.

– Section~\ref{sec:extend_vsr}  presents results of extending the proposed framework to compressed video super-resolution to demonstrate its applicability across various domains. 
 
– Section~\ref{sec:ablation} provides visual results of incorporating MV alignment and region-aware refinement, analyzing the number of experts and impact of frame adaption for improving the temporal consistency.
 
– Section~\ref{sec:details} introduces details of experimental settings, including the dataset preparation, baseline methods, and implementation details.

 – Section~\ref{sec:dicussion} discusses related works that also focus on downstream vision tasks, and further analyzes applicable scenarios of these works and the proposed method.

\section{Procedure of Quality Enhancement} \label{sec:procedure}
The goal of compressed video enhancement is to reconstruct high-quality outputs $\{\hat{y}_1, \hat{y}_2, ..., \hat{y}_T\}$ from compressed inputs $\{x_1, x_2, ..., x_T\}$. Our proposed framework achieves this through two key components: the compression-aware adaptation (CAA) network, denoted as $\mathcal{G}_{\phi}$, and the bitstream-aware enhancement (BAE) network, denoted as $\mathcal{F}_{\theta_i}$, which ensure adaptively handling different compression settings and reconstructing high-fidelity content, respectively. The overall procedure is summarized in Algorithm~\ref{algorithm}.

\noindent \textbf{Compression-aware adaptation (CAA) network $\mathcal{G}_{\phi}$}  focuses on hierarchical parameters adaptation, consisting of sequence-wise weight generator $\mathcal{G}_{\phi_s}$ and frame-wise parameters generator $\mathcal{G}_{\phi_i}$ to adaptively tailor the enhancement model to the characteristics of compressed frames (see Step~\ref{step1} and Step~\ref{step3}). The obtained frame-wise expert layer $f_{\theta_i}$ further constructs the subsequent bitstream-aware enhancement network $\mathcal{F}_{\theta_i}$ (as shown in Step~\ref{step3}).

\noindent \textbf{Bitstream-aware enhancement (BAE) network $\mathcal{F}_{\theta_i}$} frame-wisely applies techniques such as motion vector (MV) alignment  (as shown in Step~\ref{step4}) and region-aware refinement (as shown in Step~\ref{step5}) to enhance temporal consistency and reconstruct fine-detailed results.

\begin{algorithm}[t]
\caption{Procedure of Enhancing Compressed Frames}
\begin{algorithmic}[1]
\REQUIRE Sequence-wise $CRF_s$, Frame-wise $CRF_i$, Input frames \{$x_{1}$, $x_2$, ... , $x_n$\}, Motion vectors $MV$, Partition map $P_i$  \\
\ENSURE Enhanced high-quality frames \{$\hat{y}_1$, $\hat{y}_2$, ... , $\hat{y}_n$\}
 \STATE Sequence adaptation \\
 $\quad$ $f_{\theta_s}$ $\leftarrow$  $\mathcal{G}_{\phi_s}$($CRF_s$, \{$f_{\theta_1}$, $f_{\theta_2}$, ... , $f_{\theta_N}\}$)  \label{step1}

  \FOR{$x_i$ $\in$ \{$x_{1}$, $x_2$, ... , $x_T$\}}
   \STATE Frame adaptation \\ \label{step3}
  $\quad$ $\mathcal{F}_{\theta_i}$ $\leftarrow$ $f_{\theta_i}$ $\leftarrow$  $\mathcal{G}_{\phi_i}$($CRF_i$, $f_{\theta_s})$
  \STATE Motion vector alignment \\
  $\quad$ $\hat{x}_i$ $\leftarrow$ [$MV(h_i^p)$, $MV(h_i^f)$, $x_i$]  \label{step4}
  \STATE Region-aware refinement \\
  $\quad$ $\hat{y}_i$ $\leftarrow$ $\mathcal{F}_{\theta_i}$($\hat{x}_i$, $P_i$)  \label{step5}
  \ENDFOR 
  \RETURN \{$\hat{y}_1$, $\hat{y}_2$, ... , $\hat{y}_n$\}
 \end{algorithmic}   
  \label{algorithm} 
\end{algorithm}

  \begin{figure*}[t]
\centering
\includegraphics[width=1\linewidth, clip=true, trim=0 10pt 0 0]{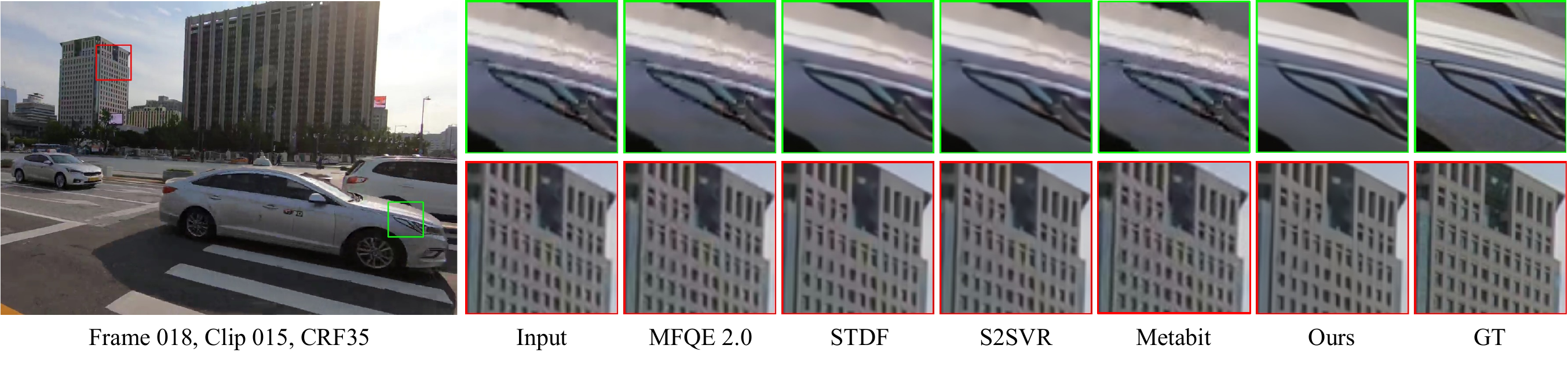} 
\includegraphics[width=1\linewidth]{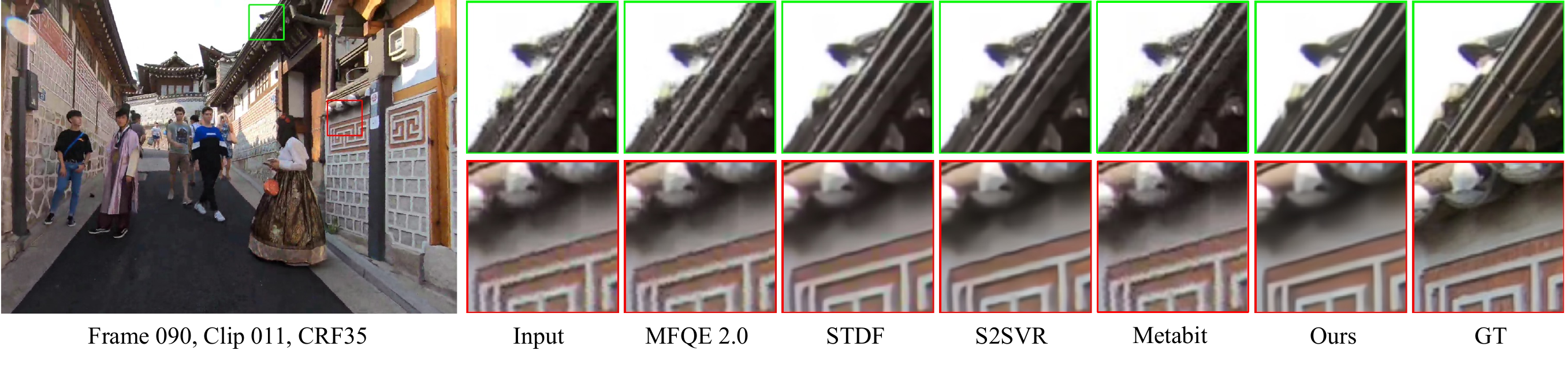}\vspace{-4pt}
\caption{Qualitative results on quality enhancement, where the results are evaluated on the  REDS4 dataset~\cite{nah2019ntire}.  As can be seen, our method demonstrates its effectiveness in reducing compression artifacts, resulting in visually appealing outputs with clear details. In contrast, the compared methods fail to fully suppress these artifacts, leaving noticeable distortions (\eg, the car in the 1st row). 
} 
\label{fig:enhance}
\end{figure*}

 \section{Quantitative Results}\label{sec:Generalization}\vspace{-3pt}  
To assess the quality enhancement performance of each method in highly compressed scenarios, we conduct evaluations at CRF values of 40, 45 and 48 and summarize the results with PSNR and SSIM (the higher the better).  Please note that the above CRF values are not included during training. The results of the REDS4 dataset~\cite{nah2019ntire} are reported in Table~\ref{tab:generalization}.    As can be seen, performing frame-wise adaptation with slice type  (marked with \colorbox{lightgrey!40}{grey})  achieves a similar performance (less than 0.03 dB in terms of PSNR) to the original design. Additionally, the proposed method shows robust performance in enhancing the highly compressed inputs, achieving PSNR gains of 0.74 dB, 0.46 dB and 0.33 dB on CRF40, CRF45 and CRF48, respectively.  In contrast, the other methods provide limited and even no improvement.  For instance, STDF~\cite{stdf} and S2SVR~\cite{lin2022unsupervised} achieve a minor PSNR gain of  0.04 dB and  0.41 dB at CRF40, respectively.   MFQE 2.0~\cite{mfqe2} and Metabit~\cite{Ehrlich_2024_WACV} show no improvement on the highly compressed inputs, indicating their dependency on a well-designed training strategy to cope with a wide range of CRFs instead of a general mix-training strategy of various compression levels.

\begin{table}[t]
\centering
\large
\resizebox{1\linewidth}{!}{
\begin{tabular}{l|c|c|c}
\toprule
\multirow{2}{*}{Method} & CRF40 & CRF45 & CRF48  \\
\cmidrule{2-4}
  & PSNR$\uparrow$  /  SSIM$\uparrow$  &  PSNR$\uparrow$  /  SSIM$\uparrow$ &  PSNR$\uparrow$  /  SSIM$\uparrow$  \\
\midrule  
Input &   26.69 / 0.7352  &   24.38 / 0.6452  &  23.17 / 0.5989   \\
 MFQE 2.0~\cite{mfqe2} &  26.69 / 0.7369  &   24.37 / 0.6466 &   23.16 / 0.6001  \\
 STDF~\cite{stdf}  & 27.03 / 0.7477  &  24.54 / 0.6544   &   23.26 / 0.6058   \\ 
  S2SVR~\cite{lin2022unsupervised} &  27.10 / 0.7506 & 24.59 / 0.6575 & 23.30 / 0.6091  \\
  Metabit~\cite{Ehrlich_2024_WACV} & 26.69 / 0.7352	& 24.38	/ 0.6452	& 23.17	/ 0.5988 \\
 \midrule  
\multirow{2}{*}{Ours}  & \cellcolor{lightgrey!40}\underline{27.42} / \underline{0.7619} & \cellcolor{lightgrey!40}\underline{24.82} / \underline{0.6697} & \cellcolor{lightgrey!40}\underline{23.47} / \underline{0.6201}  \\
& \textbf{27.43 / 0.7619} &	\textbf{24.84 / 0.6697} &	\textbf{23.50 / 0.6215} \\
\bottomrule
\end{tabular}
}\vspace{-4pt}
\caption{Quantitative results on quality enhancement, where the evaluation is conducted in highly compressed scenarios (\ie, CRF40, CRF45 and CRF48) and summarized with PSNR and SSIM (the higher the better). The best and second best results are highlighted with \textbf{bold} and \underline{underline}. Results obtained by replacing frame-wise $CRF_i$ with slice type are highlighted with \colorbox{lightgrey!40}{grey}. } 
\label{tab:generalization}
\end{table}

 \section{More Qualitative Comparisons}\label{sec:Qualitative Comparisons}
\subsection{Quality Enhancement}\label{sec:Quality Enhancement}\vspace{-2pt}
We provide visual comparisons on the task of quality enhancement in Figure~\ref{fig:enhance}. As can be seen,  MFQE 2.0~\cite{mfqe2} and Metabit~\cite{Ehrlich_2024_WACV} fail in eliminating the compression artifacts, leading to the texture distortion (\eg, the car in the 1st row). Despite STDF~\cite{stdf}  and S2SVR~\cite{lin2022unsupervised} effectively refining the compressed frames, they struggle to eliminate the color distortion and provide artifact-free results (\eg, the building in the 2nd row). In contrast, the proposed method effectively eliminates the compression artifacts and corrects the color distortion, achieving visually satisfying results.

\begin{figure*}[t]
\centering
\includegraphics[width=1\linewidth, clip=true, trim=0 10pt 0 0]{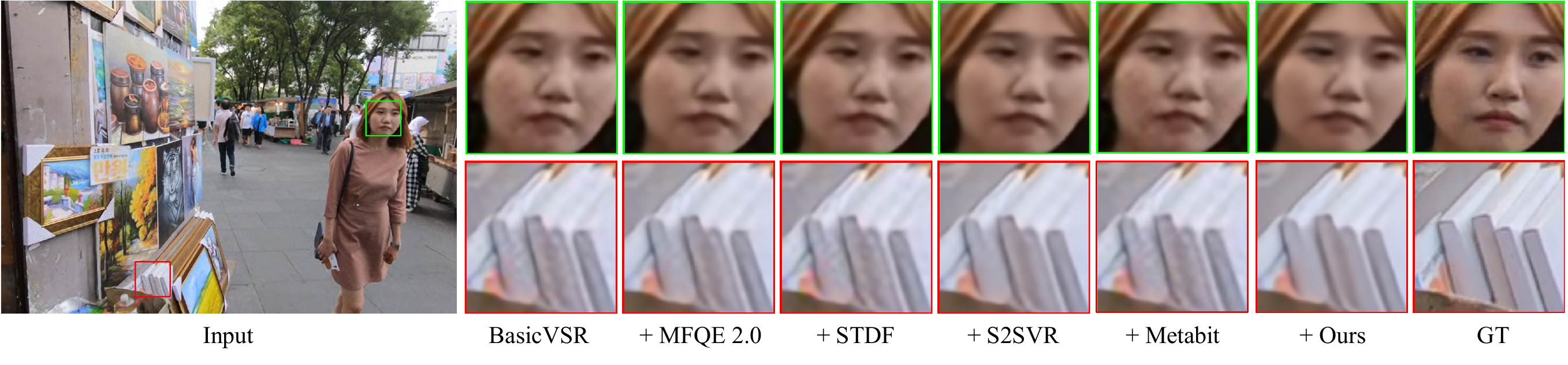} 
\includegraphics[width=1\linewidth]{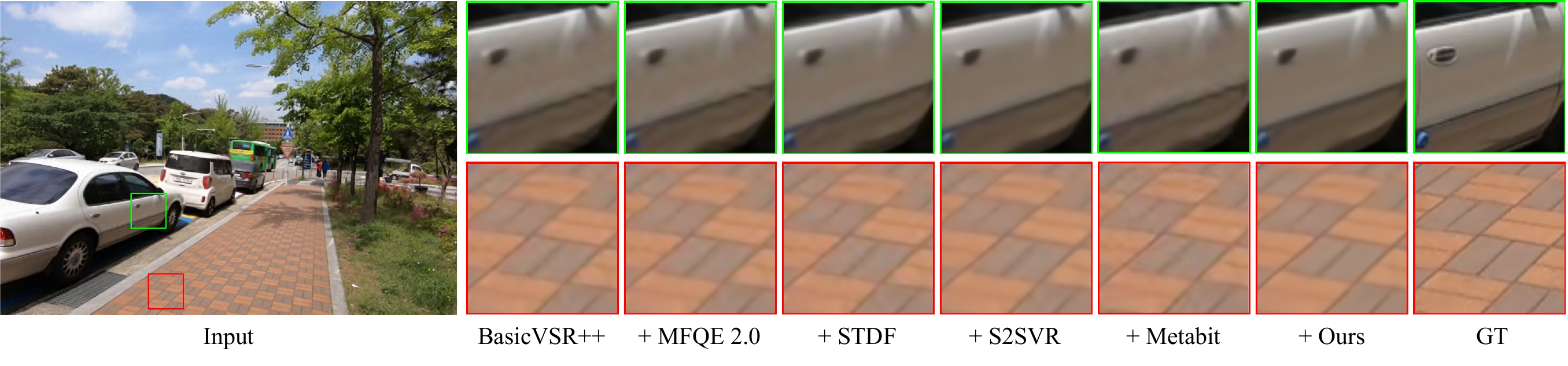}\vspace{-8pt}
\caption{Qualitative results of $\times$4 video super-resolution on the REDS4 dataset~\cite{nah2019ntire}.      As can be seen, pre-enhancing compressed frames with our method effectively prevents the amplification of compression artifacts.   While the other enhancement methods struggle to eliminate the artifacts and even severe the distortions in some cases (\eg, STDF~\cite{stdf} in the 4th row). }  
\label{fig:vsr}
\end{figure*}

\begin{figure*}[t]
\centering
\includegraphics[width=1\linewidth, clip=true, trim=0 10pt 0 0]{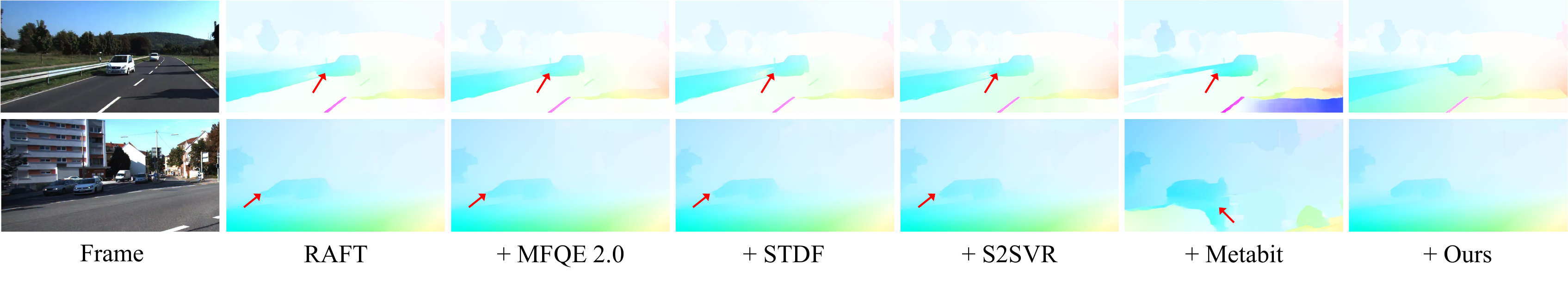} 
\includegraphics[width=1\linewidth]{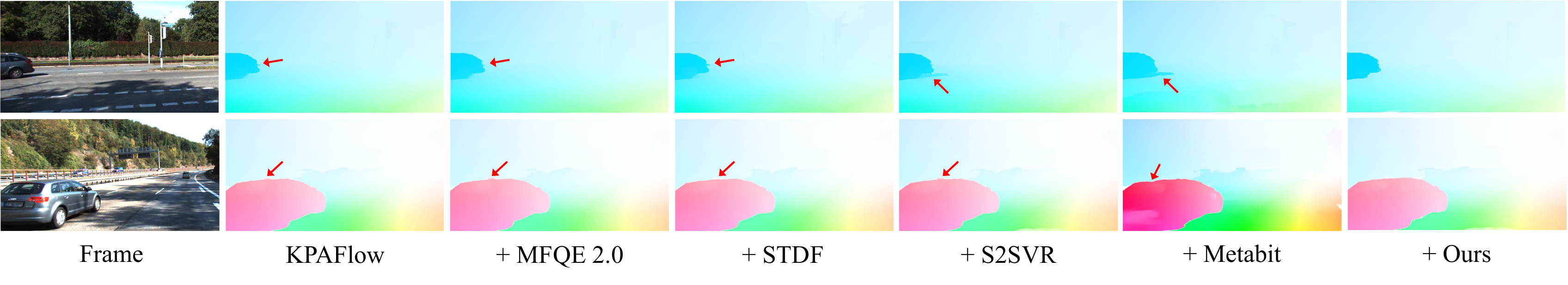}\vspace{-8pt}
\caption{Qualitative results of optical flow estimation on the KITTI-2015 dataset~\cite{geiger2013vision}, where we mark the inaccurate boundaries with red arrows.  As can be seen, equipping the baseline models with our method effectively improves the accuracy at the boundaries of moving objects (\eg, the moving car of the 1st row).  
}   
\label{fig:flow}
\end{figure*}

\subsection{Versatility Evaluation}\label{sec:Versatility Evaluation}
\noindent \textbf{Video super-resolution.} 
 As shown in Figure~\ref{fig:vsr}, it is challenging to apply video super-resolution (VSR) models that are tailored for clean data to compressed inputs, leading to the amplification of compression artifacts, as observed in the 1st column.  Equipping the baselines with pre-enhancing methods such as MFQE 2.0~\cite{mfqe2} and Metabit~\cite{Ehrlich_2024_WACV} provides limited quality improvement, and STDF~\cite{stdf} struggles to adequately suppress these artifacts (\eg, the car in the 3rd row). In contrast, pre-enhancing with our method and S2SVR~\cite{lin2022unsupervised} achieves artifact-free results, preserving the sharp edges and details of the content.  
Notably, our approach outperforms S2SVR~\cite{lin2022unsupervised}  in terms of model complexity and computational efficiency, achieving significantly lower model complexity and faster processing speeds, as detailed in Tab.~\ref{tab:QE}.

\noindent \textbf{Optical flow estimation. }  
Figure~\ref{fig:flow} presents the visualizations of predicted optical flow, with inaccurate boundaries highlighted by red arrows.   As can be seen, when estimating optical flow from compressed inputs, the inaccuracy is particularly prominent near motion boundaries  (\eg, the front of the car in the 1st row). In contrast, the proposed method demonstrates superior performance in addressing these issues, delivering more accurate results in these challenging regions compared to other methods. For instance, in the 1st row, our method effectively corrects the optical flow errors produced by RAFT~\cite{teed2020raft}, whereas both MFQE 2.0~\cite{mfqe2} and S2SVR~\cite{lin2022unsupervised} fail to provide notable improvements, and Metabit~\cite{Ehrlich_2024_WACV} perturbs the performance of downstream optical flow estimation. This highlights the effectiveness of our method in assisting the downstream optical flow estimation on compressed videos.

\begin{figure*}[t]
\centering
\includegraphics[width=1\linewidth, clip=true, trim=0 10pt 0 0]{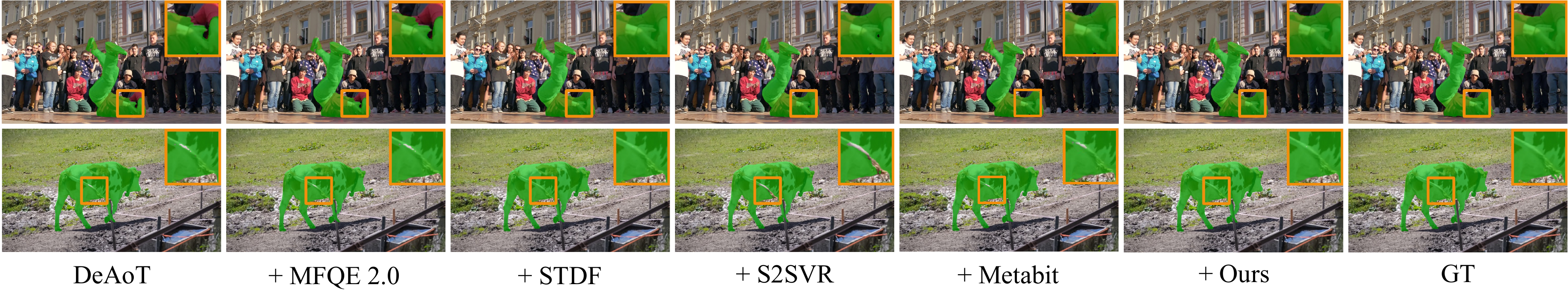}
\includegraphics[width=1\linewidth, clip=true, trim=0 10pt 0 0]{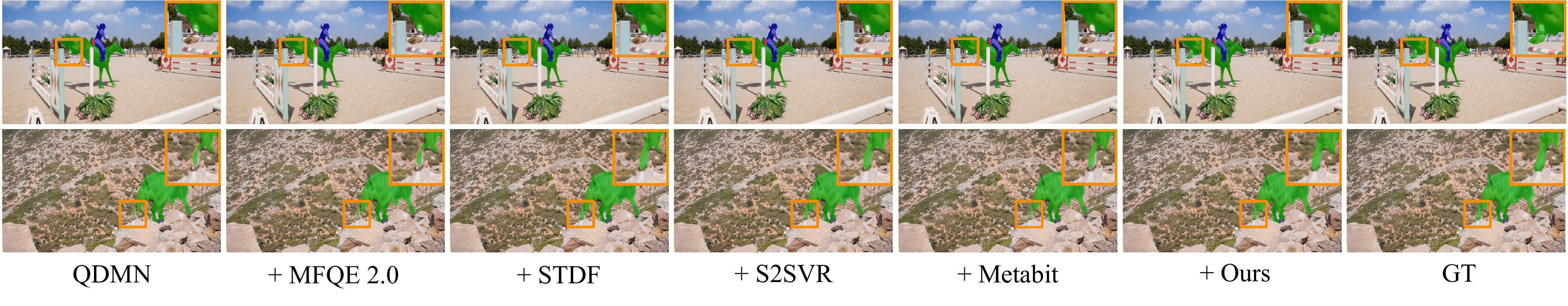}\vspace{-6pt}
\caption{Qualitative results of video object segmentation on DAVIS-17 val dataset~\cite{pont20172017}. Directly performing VOS on compressed images often results in inaccurate masks (\eg, results in the 1st column). In contrast, pre-enhancing the compressed inputs with our proposed method significantly improves mask accuracy (\eg, the tail in the 4th row).} 
\label{fig:vos} 
\end{figure*}

  \begin{figure*}[t]
    \centering  
  \includegraphics[width=1\linewidth, clip=true, trim=0 10pt 0 0]{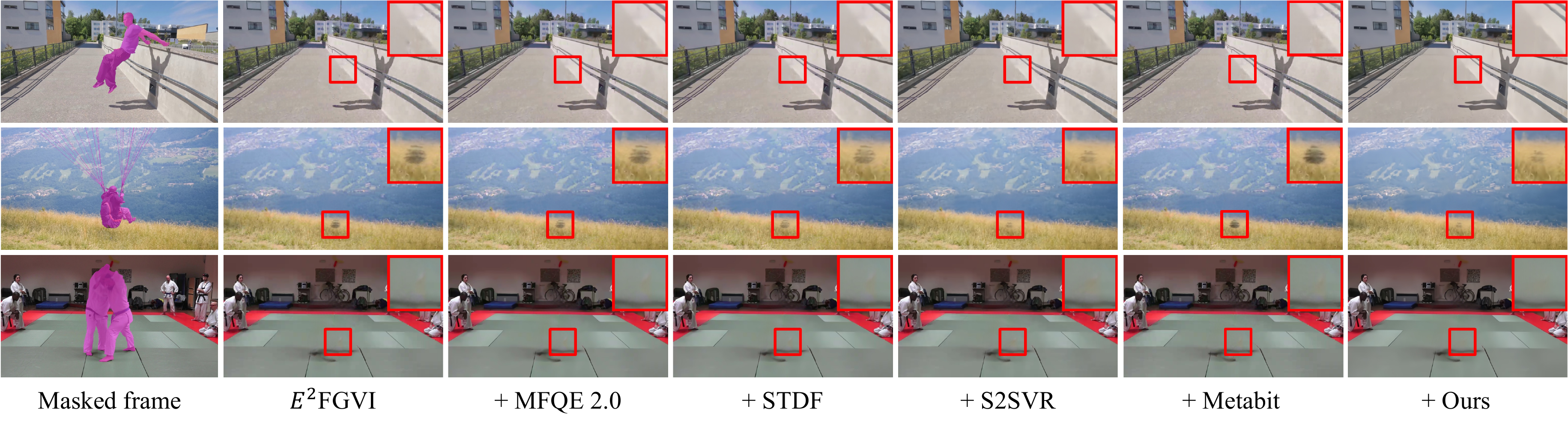}\vspace{-6pt}
    \caption{Visual results of video inpainting on the DAVIS-17 val dataset~\cite{pont20172017}. As can be seen, pre-enhancing the compressed inputs with the proposed method significantly reduces artifacts and color distortions in the removed regions (\eg, the horse hoof in the 3rd row).} 
    \label{fig:inpainting}
\end{figure*}

\noindent\textbf{Video object segmentation.} 
The results of video object segmentation are visualized in Figure~\ref{fig:vos}. As can be seen,  accurately segmenting the objects in compressed images is challenging for VOS baselines (\eg, under-segmented mask of the tail predicted by DeAoT~\cite{yang2022deaot}).  Nevertheless, such inaccuracy is not adequately -addressed by pre-enhancing the input videos with methods such as MFQE 2.0~\cite{mfqe2}, S2SVR~\cite{lin2022unsupervised}, and Metabit~\cite{Ehrlich_2024_WACV}. In contrast, the proposed method effectively mitigates errors and improves mask accuracy, underscoring the effectiveness of our method in supporting VOS on compressed video data.

\noindent\textbf{Video inpainting.}  
To further investigate the versatility of our method, we extend the downstream task to video inpainting, a generative task that needs to handle blurred object boundaries due to image compression~\cite{zhou2021deep}.  The results of removing the specified objects from compressed frames are shown in Figure~\ref{fig:inpainting}. As can be seen, due to the misalignment between compressed objects and their masks, it is hard for E$^2$FGVI~\cite{li2022towards} to adequately remove the specified object, resulting in noticeable artifacts and color distortions in the removed region (\eg, the wall in the 1st row).  In contrast, pre-enhancing the compressed inputs using our proposed method substantially improves the inpainting results, effectively mitigating artifacts and delivering results with consistent structures, demonstrating our capability of enhancing generative tasks under compression conditions.

   \begin{figure*}[t]
\centering  
\includegraphics[width=0.88\linewidth,clip=true, trim=16pt 0 54pt 0]{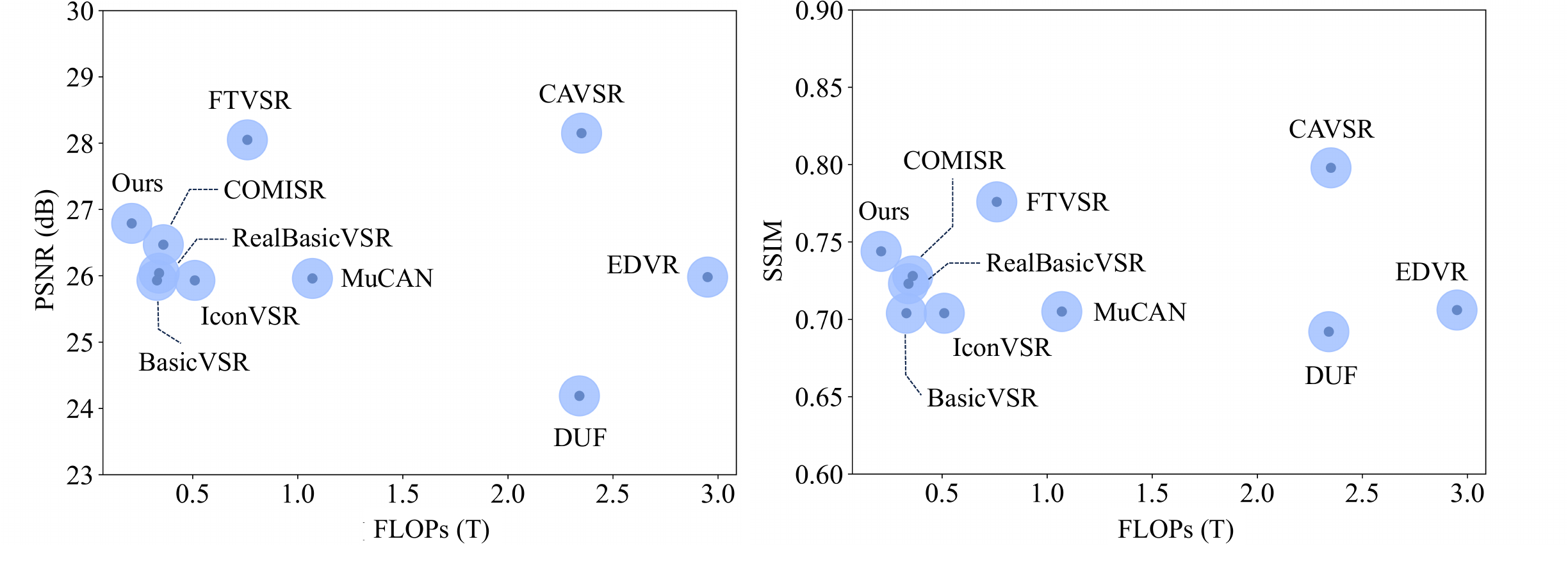}\vspace{-6pt}
\caption{FLOPs and performance comparison of $4\times$ compressed video super-resolution on the REDS4 dataset~\cite{nah2019ntire}, where the compression level is set to CRF25. Despite not being tailored for VSR, the proposed method shows competitive performance. }  
\label{fig:flops_perform}
\end{figure*}

\section{Compressed Video Super-Resolution}\label{sec:extend_vsr}
The proposed method is designed to be versatile, without any assumptions about downstream tasks, which ensures broad applicability across various domains.  Yet, it can be readily adapted for specific applications when required.  Here we demonstrate this adaptability with the application to $4\times$ video super-resolution for compressed videos. By expanding 30 region-aware refinement-integrated residual blocks and incorporating a pixel shuffle layer at the end of the network, we convert the enhancement network into a VSR-specific one. We follow COMISR~\cite{li2021comisr} to prepare the compressed training dataset and adopt the same training configuration.  The quantitative results at the compression level of CRF25 are summarized with PSNR/SSIM, and reported in Figure~\ref{fig:flops_perform}. As can be seen, although the proposed method is not tailored for VSR, it still provides competitive results with minimal computational complexity. For instance, the proposed method outperforms IconVSR~\cite{chan2021basicvsr} by 0.86 dB in terms of PSNR, costing only 0.41$\times$ of FLOPs. Additionally, our method achieves a PSNR gain over COMISR~\cite{li2021comisr} (specifically designed for compressed VSR) by 0.23 dB, while taking 0.58$\times$ FLOPs.  This indicates the versatility and potential of our method to serve as a general solution for leveraging codec information in specialized tasks.

 \begin{figure}[t]
\centering  
\includegraphics[width=1\linewidth, clip=true, trim=0 14pt 0 0]{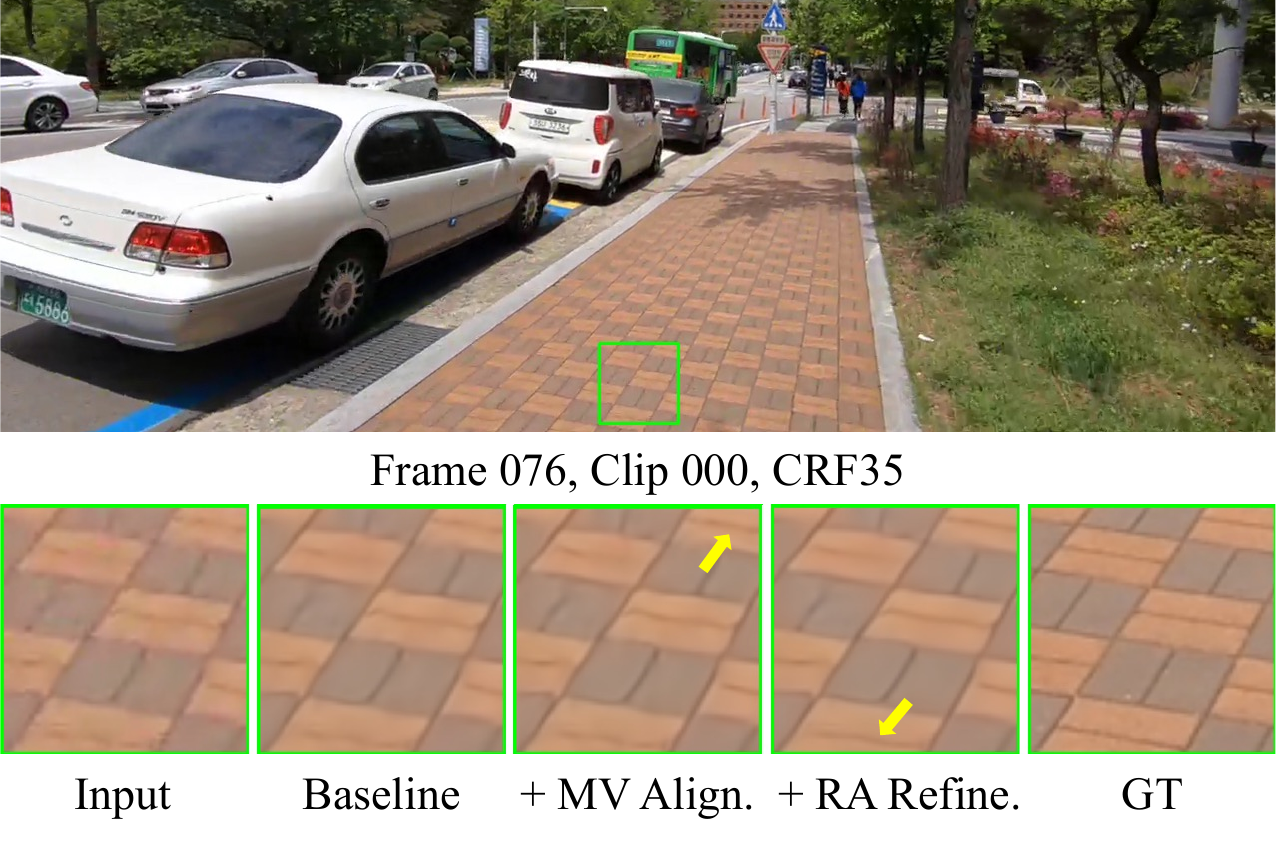}\vspace{-4pt}
\caption{Qualitative results of the ablation study on MV alignment (MV Align.) and region-aware refinement (RA Refine.). As can be seen, incorporating the region-aware refinement effectively reduces distortions and enhances the textures.}  
\label{fig:ablation_BAE}
\end{figure}

\vspace{-0.12in}\section{Ablation Studies}\label{sec:ablation}\vspace{-2pt}
In this section, we present visual results from ablation studies to assess the impact of incorporating MV alignment and region-aware refinement into the baseline model (as illustrated in Sec.~5.3 of the submission). Additionally, we analyze the effect of varying the number of experts ($N$) on model performance. These experiments are conducted on the REDS~\cite{nah2019ntire} dataset, with models trained for 50K iterations for fast evaluation. The results are summarized with PSNR and SSIM.

\noindent \textbf{MV alignment.}  
As shown in Figure~\ref{fig:ablation_BAE}, aligning frames with motion vectors (denoted as \textit{+ MV Align.}) effectively improves the texture inconsistency, as highlighted by the yellow arrow. This demonstrates the effectiveness of MV alignment in aligning and propagating high-quality reference frames, therefore improving the overall quality of compressed videos.

\noindent \textbf{Region-aware refinement. }
As shown in Figure~\ref{fig:ablation_BAE}, refining features with the guidance of partition map (denoted as \textit{+ RA Refine.}) effectively reduces distortions and enhances the fine details (\eg, the boundary of bricks marked by the yellow arrow), obtaining results with coherent textures.

\begin{figure}[t]
\hspace{-6pt}\includegraphics[width=1.05\linewidth, clip=true, trim=16pt 18pt 10pt 16pt]{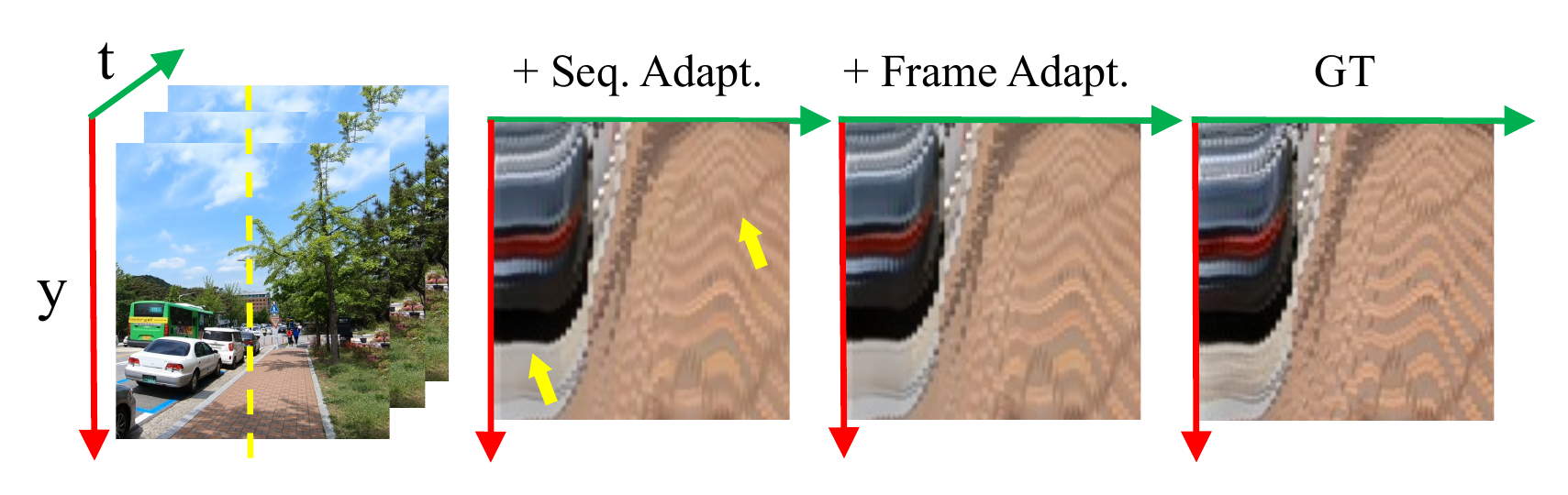}\vspace{-2pt}
\caption{Visualization of the temporal profile, which tracks a specified column (marked with the yellow dotted line) over time.} 
\label{fig:temporal}
\end{figure}

\noindent \textbf{Frame adaptation. } To assess its impact on temporal consistency, a comparison of the temporal profile is included in Figure~\ref{fig:temporal}.  As can be seen,  frame-wise adaptation helps to adaptively enhance each frame, resulting in a smoother temporal transition (as indicated by the yellow arrows).

\begin{figure}[t]
\centering  
\includegraphics[width=0.93\linewidth, clip=true, trim=24pt 32pt 20pt 10pt]{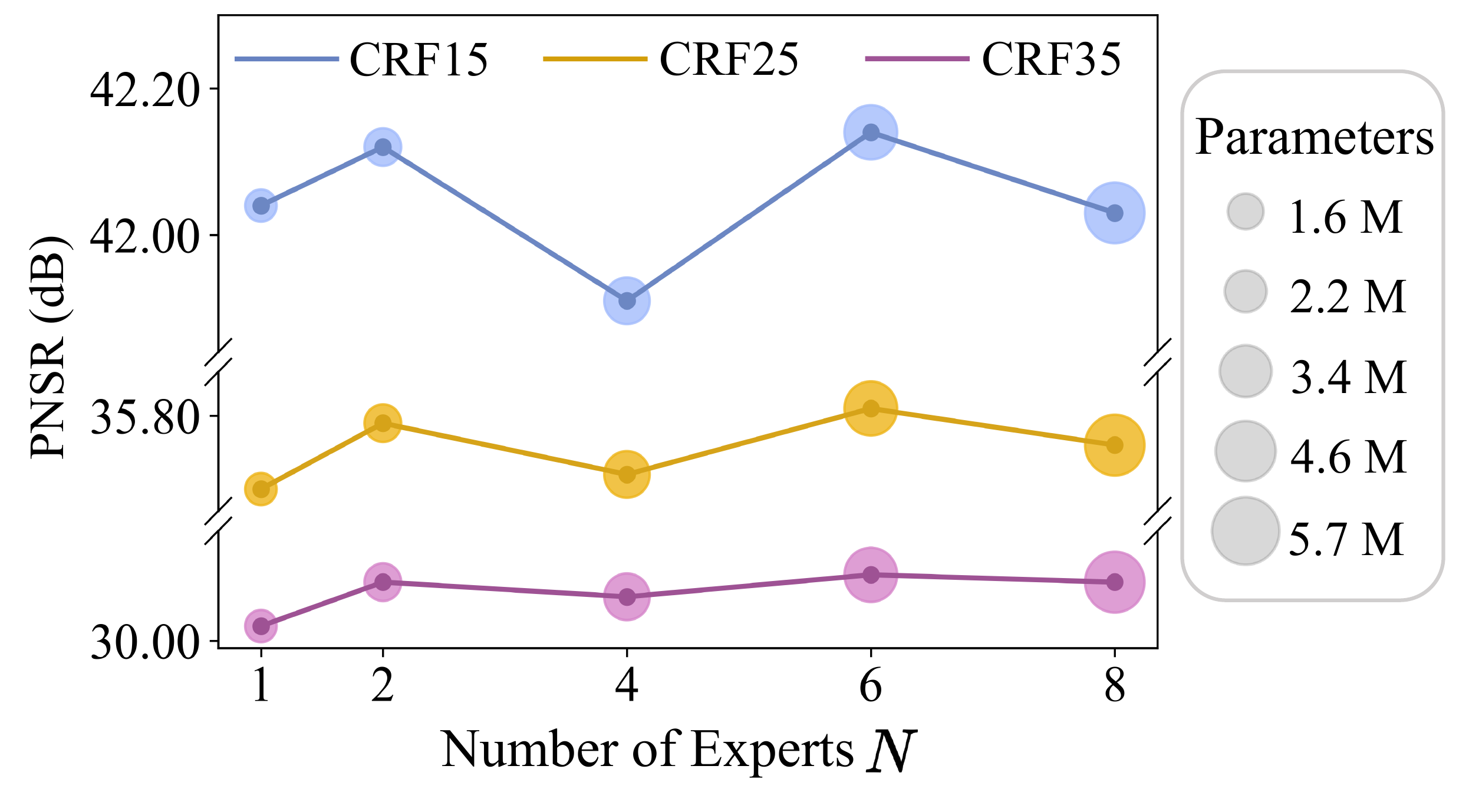}
\caption{Ablation study on the number of experts. The design of mixing experts leads to notable performance improvement, and the configuration of 6 experts is selected to balance the performance and model complexity. } 
\label{fig:expert number}
\end{figure}

\noindent \textbf{Number of experts. }
We investigate the number of experts by setting different values for $N$.  As shown in Figure~\ref{fig:expert number}, compared to a simple single-expert network, increasing $N$ effectively improves the performance but does not yield consistent performance gains.  Based on the results, we adopt $N=6$ as it achieves optimal results with manageable model complexity.

\section{Experimental Settings}\label{sec:details}
\noindent \textbf{Dataset preparation.}
We adopt the widely-used H.264~\cite{h264avc} standard and FFMPEG to generate compressed videos by specifying the CRF values (\ie, 15, 25 and 35). The $CRF_s$ value and slice type of each compressed sequence are extracted from the header. MVmed~\cite{9248145} is applied to extract motion vectors and partition maps.  
   
\noindent \textbf{Compared methods and downstream models. }
For the task of quality enhancement, we follow the official suggestions to locate keyframes with slice types for MFQE 2.0~\cite{mfqe2}.  For STDF~\cite{stdf}, we adopt the STDF-R3L variant. Since Metabit~\cite{Ehrlich_2024_WACV} only addresses I/P frames, we reimplement it to adapt the adopted dataset that contains I/P/B frames. For the task of video object segmentation (VOS), we adopt the SwinB-DeAOT-L variant from DeAoT~\cite{yang2022deaot} to ensure strong VOS performance.

 \noindent \textbf{Implementation details. }  In practice, expert layers are implemented with convolutional layers initialized with Kaiming initialization~\cite{he2015delving}. The sequence-wise weight generator is constructed with two fully connected layers followed by a softmax activation.  The parameters re-weighting is implemented with dynamic parameters mechanism~\cite{survey_dynamic}.  The frame-wise parameters generator is constructed with two fully connected layers and a sigmoid normalization. Introducing parameters $\triangle{\theta_i}$ for $f_{\theta_s}$ is implemented with dynamic transfer mechanism~\cite{se}.  The bitstream-aware enhancement network is constructed with 8 region-aware refinement-integrated residual blocks. Each block contains 64 channels.  The FLOPs and inference speed are computed with an input size of 320$\times$180 on a GeForce GTX 1080 Ti GPU. We merge the training splits of the REDS~\cite{nah2019ntire} and DAVIS~\cite{pont20172017} datasets for training, and further augment the dataset by downsampling the REDS dataset~\cite{nah2019ntire} using the Bicubic interpolation at a scaling factor of 4.    During training,  input frames are sampled from uncompressed data and compressed data with probabilities of 0.2 and 0.8, respectively. The compressed input frames 
 are sampled from CRF15, CRF25 and CRF35 with equal probability. These frames are then randomly augmented with horizontal flips, vertical flips, and rotations. The length of input sequences is set to 15 and the batchsize is set to 10.    The input patch size is set to 128$\times$128.    We adopt the Adam optimizer~\cite{kingma2014adam} with $\beta1=0.9$, $\beta2=0.99$.    The initial learning rate is set to $2\times10^{-4}$ and adjusted with the Cosine Annealing scheme~\cite{sgdr}. The whole training takes iterations of 250K. We use 2 Nvidia GeForce RTX 3090 GPUs to complete these experiments.

\section{Discussions}\label{sec:dicussion}
We explore the role of video enhancement in improving the performance of downstream tasks. Recent advancements in video codecs also introduce task-aware encoding~\cite{ge2024task} and decoding~\cite{sheng2024vnvc} frameworks to better support downstream tasks. However, these approaches typically require joint training of the compression model and target downstream tasks. In contrast, our approach serves as a plug-and-play adapter to enhance the performance of downstream models, making our method more practical, particularly in scenarios where the downstream task is unknown or subject to change.  A promising strategy would be prioritizing our approach when the downstream task is ambiguous or not specified, while leveraging the aforementioned methods when the task is well-defined and can directly benefit from the integrated task-aware compression.

\end{document}